%% file: gauss.tex
\documentclass[10pt,twoside]{amsart}

\input{preamble.tex}

\makeatletter \let\c@table\c@figure \makeatother

\usepackage{colonequals}

\usepackage[paperwidth=17cm,%
    paperheight=24cm,%
    footskip=0pt,%
    headsep=0.5cm,%
    headheight=0.5cm,%
    width=12.2cm,%
    left=2cm,%
    top=2cm,%
    bottom=2.5cm]{geometry}

\usepackage{algorithm}
\usepackage[noend]{algorithmic}

\author[K.~Heal]{Kathryn Heal}
\address{Kathryn Heal\\
School of Engineering and Applied Sciences, Harvard University, 29 Oxford Street\\ Cambridge, MA 02138, USA}
\email{kathrynheal@g.harvard.edu}
\author[A.~Kulkarni]{Avinash Kulkarni}
\address{Avinash Kulkarni\\
Department of Mathematics, Dartmouth College, Kemeny Hall, 27 N Main St\\ Hanover, NH 03755, USA}
\email{avinash.a.kulkarni@dartmouth.edu}
\author[E.~C.~Sert\"oz]{Emre Can Sert\"oz}
\address{Emre Can Sert\"oz\\
Institut f\"ur Algebraische Geometrie, Leibniz Universit\"at Hannover, Welfen-Garten 1\\ 30167 Hannover, Germany}
\email{emre@sertoz.com}

\title{Deep Learning Gauss--Manin Connections}

\date{\today}


\begin{document}

\begin{abstract}
  The Gauss--Manin connection of a family of hypersurfaces governs the change of the period matrix along the family. 
  This connection can be complicated even when the equations defining the family look simple. 
  When this is the case, it is expensive to compute the period matrices of varieties in the family via homotopy continuation.
  We train neural networks that can quickly and reliably guess the complexity of the Gauss--Manin connection of a pencil of hypersurfaces.
As an application, we compute the periods of $96\%$ of smooth quartic surfaces in projective $3$-space whose defining equation is a sum of five monomials; from the periods of these quartic surfaces we extract their Picard numbers and the endomorphism fields of their transcendental lattices. 
\end{abstract}
\maketitle

\section{Introduction}

\thispagestyle{empty}

There are two ways to study deformations of algebraic varieties. One is purely \emph{algebraic}, through the explicit polynomial equations of the family. The other is \emph{transcendental}, through the variation of Hodge structures. 
The translation of the algebraic to the transcendental is achieved through the Gauss--Manin connection associated to the family. The differential equations governing the flat sections of the connection are those that trace out the variation of Hodge structures in the corresponding flag variety.

Experimentation suggests that algebraic deformations defined by simple equations can give rise to devilishly difficult differential equations that are well beyond our ability to integrate.
In a limited capacity, this article will be concerned with the following two questions: 
Why are some Gauss--Manin connections so complicated? How can we pick deformations with milder connections?

We approach these questions from a practical point of view. 
The software presented in~\cite{sertoz18} allows us to compute the variation of Hodge structures of a pencil of hypersurfaces. 
In this article, the measure of the complexity of the Gauss--Manin connection for a pencil is the amount of time it takes to integrate its flat sections.

Our investigation led us to exploratory data analysis. We trained deep neural networks that perform well in guessing the complexity of Gauss--Manin connections of pencils of hypersurfaces, see Sections~\ref{sec:performance} and~\ref{sec:app_performance}. Their performance can be further improved by giving a local snapshot of the Gauss--Manin connection at a couple of points as in Section~\ref{sec:first_order_gm}.

Our goal in computing the variation of Hodge structures is to carry the Hodge structure (\ie periods) of one variety onto another. We determine the periods of $96\%$ of smooth quartic surfaces in $\ppp$ that can be expressed as the sum of five monomials, each with coefficient $1$. This allows us to determine their Picard number, with a small chance of error~\cite{lairez-sertoz}. 

In order to carry out a computation at this scale, we use the software in~\cite{sertoz18} as an underlying engine and add a parallelization layer, data caching mechanisms, checkpointing, and fault-tolerance. Our neural network training software is also available in this code base. We make our code available for general use\footnote{The package that brings together all the features we used is available at:\\ \url{https://github.com/a-kulkarn/period_graph}.}. 

\subsection{Pencils of hypersurfaces}\label{sec:pencils}

The Hodge structure on a smooth hypersurface $Y=Z(g) \subset \Pp^{n+1}_{\Cc}$ of degree $d$ can be represented by a matrix of periods $\cp_g \subset \Cc^{m \times m}$, where $m$ depends only on $(n,d)$, as in Section~\ref{sec:period_matrix}. The method given in~\cite{sertoz18} of computing $\cp_g$ involves deforming $g$ to another smooth hypersurface $X = Z(f) \subset \Pp^{n+1}_{\Cc}$ of degree $d$ whose periods are already known. To begin, one may take $X$ to be a Fermat type hypersurface whose periods can be expressed by closed formulas.

Given such a pair $(f,g)$ we will consider the pencil of hypersurfaces defined by $(1-t)f + tg$, which deforms $X$ to $Y$. Explicitly representing the variation of Hodge structures from $X$ to $Y$ as in Section~\ref{sec:transition}, one can compute (\ie numerically approximate) a matrix $\cp_{f,g}\in \Cc^{m \times m}$ such that if $\cp_f$ is a period matrix of $X$ then $\cp_{f,g} \cdot \cp_f$ is a period matrix of $Y$. We call such $\cp_{f,g}$ a \emph{period transition matrix}.
If $Z(h) \subset \Pp^{n+1}_{\Cc}$ is another smooth hypersurface of degree $d$. The product of the period transition matrices $\cp_{f,h}$ and $\cp_{h,g}$ gives a period transition matrix from $X$ to $Y$. 

There is a large variation on the time to compute $\cp_{f,g}$ in the inputs $f, g$. The computation of $\cp_{f,g}$ is often time consuming, taking hours or days, but for some inputs the computation of $\cp_{f,g}$ could take only a few seconds. A critical observation is that it is sometimes faster to compute $\cp_{f,h}$ $\cp_{h,g}$ than it is to compute $\cp_{f,g}$ directly.  This suggests searching for a sequence of polynomials $f{=}s_0, s_1, \dots ,s_{k} {=} g$ for which $\cp_{s_i,s_{i+1}}$ is easy to compute for each $i$. A random search based on simple heuristics was employed in~\cite[\S 3.1]{sertoz18} to find such sequences. 

Unfortunately, it is difficult to predict whether the computation of each $\cp_{s_i, s_{i+1}}$ will terminate within $k$ seconds without actually running the computation for $k$ seconds. Instead, we wish to anticipate the difficulty of such a computation so that we may discard difficult pencils in favor of friendlier ones. In this article, we approach this prediction problem with deep learning.

We cast the problem of discovering a good sequence into one of finding a short path in a weighted graph. Let $W$ be a finite set of homogeneous polynomials, all of the same degree, containing $f$ and $g$. Consider the complete graph $G$ with vertex set $W$ and some weight function $\varphi$ defined on the edges. For an edge $e$ of $G$, one may define $\varphi(e)$ to be the number of seconds it takes to compute the transition matrix~$\cp_{e}$. Our ultimate goal is to identify a path in $G$ that connects $f$ and $g$ and has small total weight. (We elaborate on how to choose $W$ in Section~\ref{sec:general_framework}.)

If the weight function $\varphi$ was known, then finding such an optimal path could be solved using standard graph traversal strategies such as Dijkstra's algorithm or the $A^*$-algorithm \cite{CherkasskyBorisV1996SpaT,RussellStuart2016Aiam}. The problem we face is that the cost of evaluating $\varphi$ at an edge $e$ is just as expensive as computing $\cp_{e}$ itself. To address this, we enlist data-driven learning models to help us guess whether $\varphi(e)$ is reasonably small or not.

We train our models on a random subset $E'$ of the edge set of the graph~$G$. We collect information on $E'$ by attempting a computation on each edge $e$ in $E'$ that is a representative fragment of the computation needed to evaluate~$\cp_e$. The models learn to recognize if, given an edge $e$ of $G$, the computation $e\mapsto \cp_e$ will terminate in a reasonable amount of time. Informed by the predictions of these models we then traverse a path in $G$ from $f$ to $g$. See Section~\ref{sec:computational_scheme} for more details on the general method and Section~\ref{sec:implementation} for our implementation of the models.

Unfortunately, the edges $e$ for which $\cp_e$ can be readily computed are generally rare. Consequently, when $k$ is a reasonably small threshold, the spanning subgraph of $G$ whose edge set is $\{e \in E(G) \mid \varphi(e) \leq k\}$ is often sparsely connected or disconnected. 
One drawback of our method is that there may not exist any good path from $f$ to $g$ inside $G$, in which case we will have wasted time trying to discover one.
Enlarging the polynomial set $W$ or the threshold $k$ may solve this problem, but at the cost of a longer training time. Further improvements may require theoretical advances into the nature of the Gauss--Manin connection. Nevertheless, what we give here significantly improves the computation time spent searching for a good connection when one exists, see~Section~\ref{sec:app_performance}.

\subsection{Deep learning in algebraic geometry}

Deep learning can be used to find elliptic fibrations~\cite{he-lee}, to recognize isomorphism classes of groups and rings~\cite{he-kim}, and for approximating the solutions of high-dimensional partial differential equations~\cite{sirignano-spiliopoulos}. Although there are symbolic algorithms for these tasks, they are impractical. Deep learning methods are employed to boost performance. The trade off for this gain in performance is the unreliability of the output. 

However, predicting the solution to a problem is not the only way one can apply deep learning methods to mathematical computations. Instead, deep learning methods can be used to assist a more reliable method by providing dynamically generated heuristics, thereby improving performance while preserving reliability -- see for instance~\cite{huang-etal}. This is our approach here, see Remark~\ref{rem:dont_approx_periods}. 

\subsection{Outline}

In Section~\ref{sec:period_transition} we give an overview of the period computation strategy for hypersurfaces. In Section~\ref{sec:computational_scheme} we explain the problem from a computational point of view and describe the resource management strategy we employ. Section~\ref{sec:neural_network} gives an overview of deep learning methodology for the non-specialist. In Section~\ref{sec:implementation} we describe our implementation of deep learning methods. In Section~\ref{sec:applications} we apply our code to five-monomial quartics and list their Picard numbers as well as their isomorphism classes.

\subsection*{Acknowledgments}

We are grateful to the following institutions for allowing us to use their computational resources: Dartmouth College,  Harvard University, Leibniz University Hannover, Max Planck Institute (MPI) MiS Leipzig. This project began while all three authors were at MPI MiS, we thank this institution for providing a stimulating environment. In addition, we thank Pierre Lairez for helpful comments. 

Avinash Kulkarni has been partially supported by the Simons Collaboration on Arithmetic Geometry, Number Theory, and Computation (Simons Foundation grant 550033) and by the Forschungsinitiative on Symbolic Tools at TU Kaiserslautern during the course of this project.

\section{Period computation}\label{sec:period_transition}

Let $X=Z(f_X) \subset \Pp^{n+1}_{\Cc}$ be a smooth hypersurface where $f_X \subset \Cc[x_0,\dots,x_{n+1}]$ is a degree $d$ homogeneous polynomial. By the Lefschetz hyperplane theorem all cohomology groups of $X$ are trivial (either $0$ or $\Zz$), except for the middle (singular) cohomology group $\H^n(X,\Zz)$. 

We need to represent two kinds of structure on the cohomology groups: the integral structure and the Hodge decomposition on $\H^n(X,\Cc)$. We recall their definition and summarize their method of computation here. For an exhaustive account, see~\cite{voisin-2007-volI, voisin-2007-volII}. The definitions that are more specific to this article are introduced in Section~\ref{sec:transition}.

\subsection{The integral structure}

As an abstract group, $\H^n(X,\Zz)$ is isomorphic to $\Zz^m$ for some $m$. After choosing an identification $\psi \colon \H^n(X,\Zz) \isoto \Zz^m$ the intersection product on $\H^n(X,\Zz)$ can be represented by an $m\times m$ matrix $\ci$ with integral entries. An integral structure on $\H^n(X,\Cc) = \H^n(X,\Zz) \otimes_{\Zz} \Cc$ refers to an identification of the sublattice $\H^n(X,\Zz) \subset \H^n(X,\Cc)$.

Although we will suppress this from notation, whenever we refer to a trivialization $\psi$ of the integral cohomology, we also mean a determination of the intersection product $\ci$ on $\Zz^m$. The integral structure on $\H^n(X,\Cc)$ can be represented by a map $\psi_\Cc \colonequals \psi \otimes \Cc$.

\subsection{The Hodge structure on cohomology}

The Hodge decomposition on $\H^n(X,\Cc) = \H^n(X,\Zz)\otimes_{\Zz} \Cc \simeq \Cc^m$ is a direct sum decomposition:

\begin{equation}
  \H^n(X,\Cc) = \bigoplus_{p=0}^n \H^{p,n-p}(X),
\end{equation}
where $\H^{p,q}(X)$ is the space of $(p,q)$-forms. 

The Hodge pieces $\H^{p,q}(X)$ do not vary holomorphically in $X$. Therefore, it is more natural for variational problems to consider the \emph{Hodge filtration} $F^{\ell}(X) = \bigoplus_{p \ge \ell}^n \H^{p,n-p}$ for $\ell = 0,\dots,n$. 
Of course, on an individual hypersurface, one can recover the decomposition from the filtration and vice versa.

Using a generic hyperplane section of $X$, we can define the hyperplane class $h \in \H^2(X,\Zz)$ in cohomology. If $n = \dim X$ is even then $h^{n/2} \in \H^n(X,\Zz)$ is called a polarization. In this case, the primitive part of the cohomology $\H^n(X,\K)_\prim$ is the orthogonal complement of $h^{n/2}$, where $\K$ is any ring. If $n$ is odd then we set $\H^n(X,\K) = \H^n(X,\K)_\prim$. Let $m_\prim \colonequals \dim_{\Cc} \H^n(X,\Cc)_{\prim}$. The restrictions $F^\ell(X) \cap \H^n(X,\Cc)_{\prim}$ of the Hodge filtrations to the primitive cohomology will be denoted by $F^{\ell}(X)_{\prim}$.

\subsection{The period matrix}\label{sec:period_matrix}

Both the integral structure and the Hodge structure are discrete invariants in isolation. 
After all, we know $\H^n(X,\Zz) \simeq \Zz^m$ and we know the dimensions of the pieces of the Hodge decomposition~(e.g.~\cite[\S 17.3]{arapura--algebraic_geometry}). The difficulty is putting these pieces together, which requires transcendental invariants --- the periods of $X$.

\begin{definition}\label{def:periods}
  Let us call $\cp \in \Cc^{m\times m}$ a \emph{period matrix} on $X$ if there is an isomorphism $\psi \colon \H^n(X,\Zz) \simeq \Zz^m$ such that for each $\ell=0,\dots,n$, the first $\dim_\Cc F^{\ell}(X)$ rows of $\cp$ span $\psi_\Cc \left( F^{\ell}(X) \right) \subset \Cc^m$.  Similarly, we define a \emph{primitive period matrix} $\cp_\prim \in \Cc^{m_{\prim} \times m_{\prim}}$.
\end{definition}

\begin{remark}\label{rem:extend_periods}
  It is clear that a period matrix $\cp$ can be obtained from the primitive period matrix $\cp_{\prim}$ and vice versa. See~\cite[\S 7]{lairez-sertoz} for an explicit computation.
\end{remark}

\subsection{The Griffiths basis for cohomology}

Different identifications of $\H^n(X,\Zz)$ with $\Zz^m$ will change $\cp$ by the action of the discrete group $\gl(m,\Zz)$. However, there is a continuous family of choices to be made in choosing a basis for $F^\ell(X)$. For a hypersurface $X$, one can resolve the latter indeterminacy by specifying a well-defined basis for the filtration. 

Let $S=\Cc[x_0,\dots,x_{n+1}]$, $\jac(f_X) = (\del_0 f_X,\dots,\del_{n+1} f_X)$ be the Jacobian ideal and $R=S/\jac(f_X)$. Since $f_X$ is smooth, $R$ is a finite dimensional algebra over $\Cc$. Let us write $R_{\ell}$ for the quotient of the homogeneous part $S_{\ell}/\jac(f_X)_{\ell}$. 

For each $\ell \ge 0$, Griffiths~\cite{griffiths--periods} defines a \emph{residue map}:
\begin{equation}\label{eq:R_to_G}
  S_{(n+1-\ell)d -n-2} \to F^{\ell}(X)_\prim : p \mapsto \frac{p}{f^{n+1-\ell}}\vol_{\Pp^{n+1}_\Cc},
\end{equation}
where $\vol_{\Pp^{n+1}_\Cc}$ is a natural generator of the twisted canonical bundle of $\Pp^{n+1}$, namely $\Omega^{n+1}_{\Pp^{n+1}/\Cc}(n+2)$, and is given by
\begin{equation}
  \vol_{\Pp^{n+1}_\Cc} \colonequals \sum_{i=0}^{n+1} (-1)^{i} x_i \dd x_0 \wedge \dots \wedge \widehat{\dd x_i} \wedge \dots \wedge \dd x_{n+1}.
\end{equation}
These residue maps descend to an isomorphism on the quotients:
\[
  \res \colon R_{(n+1-\ell)d -n-2} \isoto F^\ell(X)_\prim/F^{\ell+1}(X)_{\prim}, \quad \forall \ell=0,\dots,n.
\]

Impose the grevlex ordering on $S$ and consider the ideal of leading terms $\lt(\jac(f_X))$ of $\jac(f_X)$. The grevlex ordering gives a well-defined sequence of monomials which descend to a basis of $S/\lt(\jac(f_X))$, and therefore to a basis of $R$. Their residues $\omega_1,\dots,\omega_{m'}$ in appropriate degrees yields a basis of the primitive cohomology $\H^n(X,\Cc)_{0}$. 

\begin{definition}
  The basis $\omega_1,\dots,\omega_{m'}$ of the primitive cohomology constructed above is a well-defined basis which respects the filtration. We will call this basis \emph{the (grevlex) Griffiths basis} on $X$.  
\end{definition}

\begin{definition}
  A primitive period matrix $\cp_\prim$ as in Definition~\ref{def:periods} will be called a \emph{primitive grevlex period matrix} of $X=Z(f_X)$ if the $i$-th row of $\cp_\prim$ equals $\psi_{\prim,\Cc}(\omega_i)$, where $\psi_{\prim} \colon \H^n(X,\Zz)_{\prim} \isoto \Zz^{m_\prim}$ and $\{\omega_j\}_{j=1}^{m_\prim}$ is the grevlex Griffiths basis on $X$. Any period matrix obtained by extending $\cp_{\prim}$ as in Remark~\ref{rem:extend_periods} will be called a \emph{grevlex period matrix} of $X$.
\end{definition}

Any two grevlex period matrices of $X$ differ by the action of the discrete group $\gl(m,\Zz)$. Moreover, after fixing a coordinate system on the integral primitive cohomology, the grevlex period matrix is uniquely defined. 

\subsection{Transition matrices for periods}\label{sec:transition}

Consider a pair of smooth hypersurfaces $X=Z(f)$ and $Y=Z(g)$ with $f,g \in \Cc[x_0,\dots,x_{n+1}]_d$. 

\begin{definition}
  We will call a matrix $\cp_{f,g} \in \Cc^{m_{\prim}\times m_{\prim}}$ a \emph{period transition matrix} if for any grevlex primitive period matrix $\cp_f$ on $X=Z(f)$, the product $\cp_{f,g} \cdot \cp_f$ is a primitive grevlex period matrix on $Y=Z(g)$.
\end{definition}

A method for computing a period transition matrix $\cp_{f,g}$ is explained in~\cite{sertoz18}. A much simpler method is available when $g$ is a linear translate of $f$ and we will do this example in Section~\ref{sec:linear_translate}. We will briefly outline the general method of~\cite{sertoz18} here.

\begin{enumerate}
  \item Take a one parameter family of hypersurfaces $\cx_t = Z(f_t)$ where $f_t \in \Cc(t)[x_0,\dots,x_{n+1}]_d$ with $X=\cx_0$ and $Y=\cx_1$. 
  \item Find polynomials $p_1,\dots,p_{m'} \in \Cc(t)[x_0,\dots,x_n]$ which descend to a basis for $\bigoplus_{\ell=0}^{n} \Cc(t)[x_0,\dots,x_n]_{(n+1-\ell)d -n-2}/\jac(f_t)_{(n+1-\ell)d -n-2}$ and also to bases when $t=0$ and $t=1$. Often $p_i$ are monomials or a sum of two monomials (with constant coefficient $1$).
  \item Find the matrix $B$ expressing the change of basis from the basis above to the grevlex basis at $t=1$. This is done by the computation of normal forms.
  \item For each $i=1,\dots,m'$ find a differential operator $\cd_i \in \Cc(t)[\frac{\del}{\del t}]$ such that $\cd_i \cdot \res(p_i) = 0$. These differential equations annihilate the rows of the grevlex period matrix of $\cx_t$.\label{item:find_cd}
  \item Find a path $\gamma$ in $\Cc \setminus \cs$ where $\cs$ is the set of values of $t$ for which $\cx_t$ is singular. 
  \item Compute the transition matrix for the space of solutions of $\cd_i$ at $t=0$ and $t=1$ obtained by homotopy continuation along $\gamma$.\label{item:integrate_cd}
  \item Using the indicial equation of $\cd_i$, find the set of derivatives of $\res(p_i)$ whose values at $t=0$ would determine $\res(p_i)$. 
  \item Express the derivatives of $\res(p_i)$ at $t=0$ in terms of the grevlex Griffiths basis at $t=0$. Multiply this expression on the right with the transition matrix of \eqref{item:integrate_cd} and take the first row.
  \item Form the $m' \times m'$ matrix whose $i$-th row is the row obtained at the step above and multiply it by $B$ on the left to get $\cp_{0,1}$.
\end{enumerate}

The most time expensive step is step~\eqref{item:integrate_cd}. However, most attempts to compute the period transition matrix fail on step~\eqref{item:find_cd}. This is because it requires a Gr\"obner basis computation for $\jac(f_t)$, numerous normal form computations, and expressions of elements in $\jac(f_t)$ in terms of the generators $(\del_0 f_t,\dots,\del_{n+1} f_t)$.

\subsubsection{Period transition matrices of linear translates}\label{sec:linear_translate}

If two hypersurfaces are linear translates of one another, then a period transition matrix between them can be easily computed. We will do this here as an instructive example. This is also a computation we will use later in the article in Section~\ref{sec:applications}.

The general linear group $\gl(n+1,\Cc)$ acts on the coordinates on $\Pp^{n+1}$ linearly. The induced action on the coordinate ring $S=\Cc[x_0,\dots,x_{n+1}]$ is given by $u \cdot f(x) = f(x \cdot u^t)$ where $u \in \gl(n+1,\Cc)$, $f \in S$, $u^t$ is the transpose of the matrix $u$, and $x=(x_0,\dots,x_{n+1})$ is treated as a row vector. 

With $X=Z(f)$ a smooth hypersurface as before, suppose that $Y=Z(g)$ with $g=u\cdot f$ for some $u \in \gl(n+1,\Cc)$. If $\cp_X$ is a period matrix of $X$ then it certainly works as a period matrix for $Y$. However, even if $\cp_X$ is a grevlex period matrix on $X$ it need not be grevlex on $Y$. We describe the computation of a period transition matrix $\cp_{f,g}$ below.

Let $p_1,\dots,p_{m'} \in S$ be the polynomials whose residues give the grevlex Griffiths basis on $X$. Let $\phi\colon Y \isoto X$ be the isomorphism induced by $u$. If $\cp_X$ is a primitive grevlex period matrix on $X$ then $\cp_X$ is a primitive period matrix on $Y$ whose rows represent the residues of $u\cdot p_1, \dots , u\cdot p_{m'}$. 

We can use Griffiths--Dwork reduction (see \cite{griffiths--periods} or a summary \cite[\S 2.4]{sertoz18}) on $Y$ to write each $\res(u\cdot p_i )$ in terms of the grevlex Griffith basis on $Y$. If $N$ is a matrix whose rows store the coordinates of $\res(u \cdot p_i)$ in the grevlex basis then $N\inv \cdot \cp_X$ will be a primitive grevlex period matrix on $Y$. In other words, $N\inv$ is a period translation matrix from $X$ to $Y$. This computation is implemented in {\tt PeriodSuite}\footnote{https://github.com/emresertoz/PeriodSuite} as the function {\tt translate\_period\_matrix}.

\subsubsection{First order Gauss--Manin connection}\label{sec:first_order_gm}

We would like to detect which of the differential operators $\cd_i$ appearing in Item~\eqref{item:find_cd} of Section~\ref{sec:transition} would be easy to integrate, before we compute $\cd_i$. 

Collectively, these $\cd_i$ define flat sections of the Gauss--Manin connection of the family $\cx_t$~(see for example~\cite[\S 9.3]{voisin-2007-volI}). Computing the Gauss--Manin connection itself is simpler than computing its flat sections, but still not quite fast enough for rapid testing. Computing the Gauss--Manning connection evaluated at a single point, however, is very fast. Furthermore, these evaluations give an impression of how complicated $\cd_i$ might be (see Section~\ref{sec:learning_computability}).

Let $\{\omega_i = \res(p_i)\}_{i=1}^{m'}$ be a basis for the primitive cohomology on a hypersurface $X$. Say $X$ is the fiber of a family $\cx_t$ at $t=t_0$. Then the residue of the polynomials $p_i$ on the family $\cx_t$ will give a basis for the primitive cohomology for all $t$ near $t=t_0$. Differentiating these forms with respect to $t$ and evaluating at $t=t_0$ gives new elements in the cohomology of $X$. Expressing these new forms in terms of $\{\omega_i\}_{i=1}^{m'}$ gives an $m' \times m'$ matrix. We call this matrix the \emph{first order Gauss--Manin matrix}. 

Let us note that if $X$ and the polynomials $p_i$ are defined over a subfield $K \subset \Cc$ then the corresponding first order Gauss--Manin matrix will have entries in $K$. See~\cite{katz-oda} for more on this topic. In our applications, we will take $K=\Qq$.

\section{Computational scheme}\label{sec:computational_scheme}

We will now describe our approach to the problem of searching for a good path between polynomials, with the goal of transferring their periods from one to the other. There are two variations of this problem that we are interested in solving. We will state and explain these variations and then generalize them to a common framework in Section~\ref{sec:problem_types}. In Section~\ref{sec:search_computation_graph} we will explain how we operate in this abstract framework.

\subsection{Two types of problems and a general framework}\label{sec:problem_types}

We are interested in solving two types of problems. In the first problem, we are given a pair of polynomials $V=\{f,g\}$ depicting smooth hypersurfaces of the same degree, and our goal is to compute the period transition matrix $\cp_{f,g}$.  In the second problem, we are given a (possibly large) set of polynomials $V$, and our goal is to compute the periods of all elements in $V$ given the periods of any one of them. Let us present some example problems we will solve later on.

\subsubsection{First problem: computing periods for one target polynomial}\label{sec:first_problem}

For a general $(f,g)$, a direct computation of $\cp_{f,g}$ is infeasible even with low precision. The path from $f$ to $g$ must be broken into simpler pieces. The strategy we discussed in Section~\ref{sec:pencils} is to find a sequence of polynomials $f=h_0,h_1,\dots,h_s=g$ such that each intermediate period transition matrix $\cp_{h_i,h_{i+1}}$ is easily computable. Their product would then give $\cp_{f,g}$. This problem was already investigated in~\cite[\S 3.1]{sertoz18} and a crude heuristic was developed there. We will develop this heuristic further in this section. 

\subsubsection{Second problem: computing periods for many hypersurfaces}\label{sec:second_problem}

Here we are given a (possibly large) set of polynomials $V$ and we would like to be able to compute the periods of all elements in $V$ given the periods of any one of them. For example, the set $V$ may be the set $V_n$ consisting of all smooth quaternary quartics that are expressed as the sum of $n$ distinct monomials with coefficients equal to $1$, e.g.
\[
  V_4 = \{x^4+y^4+z^4+w^4,\, x^3y + xy^3 + z^3w + w^4, \dots\}.
\]
An attempt to compute the periods, and therefore the Picard numbers, of all elements in $V_5$ was made in~\cite{lairez-sertoz}. Many of the elements in $V_5$ were out of reach at the time. We apply our methods to compute the periods of most elements in $V_5$, see Section~\ref{sec:applications}.

Given a sequence of polynomials $f=h_0,h_1,\dots,h_s=g$ such that each intermediate period transition matrix $\cp_{h_i,h_{i+1}}$ is easily computable, and such that the periods of $f$ are known to some precision, we easily obtain the periods of each intermediate quartic by the partial products $\cp_{h_{i}, h_{i+1}} \ldots \cp_{h_{0}, h_{1}} \cp_f$. Thus, it suffices to determine a set of paths which connect the vertices of $V$ as opposed to computing the periods one by one.

\subsubsection{General framework}\label{sec:general_framework}

Both problems can be slightly generalized to fit into the following framework. Suppose we are given a set $V$ containing one element whose periods are known and the periods of all the others are sought.

The set $V$ may not have enough pairs $f,g \in V$ such that $\cp_{f,g}$ is directly computable. In this case, we need to construct a larger set $W$ containing $V$ which introduces many pairs $(f,g) \in V \times W$ such that $\cp_{f,g}$ is directly computable. The construction of this $W$ can be based on human heuristics.

Letting $K_W$ be the complete graph with vertex set $W$, we now wish to solve the following problem:
{\em
  Given $V \subset W$, find a tree $T \subset K_W$ such that the vertex set of $T$ contains $V$ and the computation of the period transition matrix for each edge in $T$ is feasible.
}
Constructing $W$ is a balancing act. If the search space $W$ is too large, it may be impractical to \emph{find} a good tree $T$ inside $K_W$, even though one \emph{may exist}. Conversely, if $W$ is too small, we may be able to search the entire space but find that there is no good tree $T$ inside $K_W$. 

For our first type of problem (Section~\ref{sec:first_problem}), with $V=\{f,g\}$, one may take the following set:
\[
  W = \{h \mid \supp(h) \subset \supp(f)\cup \supp(g),\, \coefs(h) \subset \coefs(f) \cup \coefs(g)\}
\]
where $\supp$ denotes the monomial support of a polynomial and $\coefs$ denotes the set of coefficients of a polynomial.
For the second type of problem we will consider $V = V_5$. In this case, it seems natural to take something in between $W=V_4 \cup V_5$ and $W=V_4 \cup V_5 \cup V_6$.  

\begin{remark}
  There are numerous modifications to this problem that can help attain the final goal. We will only point out one: it is often sensible to prune $K_W$ by a problem specific heuristic before embarking on a search for $T$. 
\end{remark}

\subsection{Searching a computation graph for an efficient tree}\label{sec:search_computation_graph}

In this section, we no longer discuss period computations. We are interested only in abstracting the problem laid out in Section~\ref{sec:general_framework} to, essentially, a problem in resource management in a computational exploration.

Let $G=(W,E)$ be a graph with vertex set $W$ and edge set $E$. Consider a computationally expensive program $\mathcal{P}$ which takes as input an edge $e\in E$ and returns an output $\cp_e$. We will consider a cost function $\varphi \colon E \to \Rr_{+}$ that measures the difficulty of performing the computation $e \mapsto \cp_e$. Common examples of $\varphi(e)$ would be the number of arithmetic operations needed to complete the program $e \mapsto \cp_e$ or the amount of memory required to execute the program. For our purposes, it is important that a lower bound for $\varphi$ can be determined in finite time. Henceforth, we fix one such $\varphi$ and refer to it as a \emph{complexity function}. 

\begin{remark}
  In the context of this paper, $W$ is a set of polynomials. For each edge $e$, our procedure outputs a period transition matrix $\cp_e$ to a fixed degree of precision. The value of $\varphi(e)$ may then be the number of seconds it took to perform the computation $e \mapsto \cp_e$ on a fixed computer. These specifics will not be relevant for the rest of this section, as we continue with the abstraction above. 
\end{remark}

One could rephrase the problem introduced in the previous section as follows: 
\emph{Given $V \subset W$, find a tree $T \subset G$ such that $V$ is contained in the vertex set of $T$ and $\varphi(T) = \sum_{e \in T} \varphi(e)$ is small, if not minimal. }

If we know the value of $\varphi$ on each edge, then this problem becomes a standard minimization problem. However, at the outset, the evaluation of $\varphi$ is just as expensive as the computation $e \mapsto \cp_e$ itself. 
This leads to the following brute-force strategy.

\subsubsection{Brute-force strategy}

    \begin{algorithm}
	\caption{Brute force with thresholding} \label{algo:brute_force}
	\begin{algorithmic}[1]
    \REQUIRE $ $ \\
		$V$ -- Set of target vertices \\
		$W$ -- Set of waypoint vertices containing $V$ \\
		$E$ -- List of edges to be attempted \\
		$k$ -- Threshold parameter, e.g. maximum time per attempt \\
		$Q$ -- The job queue, \ie a bijection $Q \colon \{1, \dots, \#E \} \to E$

		\ \\
		
		\STATE
		Let $G'=(W,\emptyset)$ be the graph on $W$ with empty edge set.
		
		\WHILE{$n \le \# E$}

		\STATE
			\textbf{Attempt} the computation $n \mapsto \cp_{Q(n)}$. \\ {}
			\textbf{Abort} if $\varphi(Q(n)) \geq k$ is detected.

		\IF {successful}
			\STATE
			Set $G' := G' \cup \{e\}$
			
			\IF {$V$ is contained in a connected component of $G'$}
				\RETURN $G'$
			\ENDIF
		\ENDIF
		
		\STATE
		Increment $n$ \\

		\ENDWHILE
		
		\RETURN Fail
		
		\COMMENT{(One may apply multiple rounds with enlarged $W$ and $k$ to eventually succeed.)}	
	\end{algorithmic}
\end{algorithm}

The brute force method (Algorithm~\ref{algo:brute_force}) attempts to compute every edge in the graph until the desired tree is constructed. If the computation of an edge takes too long, then the computation is aborted and the computation for the next edge is begun.

  In practice, we perform the computations in the job queue in parallel, but the size of the edge set ($\approx 10^6$) is so much larger than the number of cores available to us ($\approx 10^2$) that this serial conceptualization is not too far off the mark.

\subsubsection{Informed brute force}\label{sec:informed_brute}

It is clear that the choice of the ordering $Q$ on the edge set will make a dramatic impact on the total time it takes to find $T$. In ideal cases, the first few edges in the sequence might be easily computable and sufficient to form a good tree $T$. We could then stop searching early. In unfortunate cases, all of the edges $e$ for which $\varphi(e)>k$ might be queued up first in the list, in which case we would end up spending all our time trying to compute impossible edges. For this reason, we set finding a good ordering $Q$ as our top priority.

A strategic choice of a statistical model will be used to give us a function $\phi \colon E \to [0,1]$ such that if the \emph{computability score} $\phi(e)$ is close to $1$ then $\varphi(e)$ is  likely at most $k$. Moreover, as we perform computations, we can refine our model in order to improve the reliability of this function $\phi$. This suggests the modification of the brute force algorithm given by Algorithm~\ref{algo:modified_brute_force}.

    \begin{algorithm}
	\caption{Modified brute force} \label{algo:modified_brute_force}
	\begin{algorithmic}[1]
		\REQUIRE  $ $ \\
		$V$ -- Set of target vertices \\
		$W$ -- Set of waypoint vertices containing $V$ \\
		$E$ -- List of edges to be attempted \\
		$k$ -- Threshold parameter. (Maximum time per attempt, in seconds.) \\
		$\phi$ -- [Optional] Heuristic approximation to the cost function

		\ \\
		
		\IF {$\phi$ is provided}
		
			\STATE
			Set $\leq\!(e,f) := (\phi(e) \leq \phi(f))$ \ \ (function returning a Boolean)
			
			\STATE
			Set $Q := \mathtt{sort\_descending\_order}(E, \leq)$

		\ELSE
			\STATE
			Set $Q := \mathtt{randomize\_order}(E)$
		\ENDIF
				
		\STATE
		Set $\text{result} := \mathtt{BruteForce}(E, k, Q)$. \\
		Collect data during this computation in order to retrain the models and improve the reliability of $\phi(e)$.

		\IF {result is not Fail}
			\RETURN result
		\ENDIF
		
		\STATE
		Possibly enlarge $W$ or increase $k$.

    \STATE
    Retrain $\phi$.
		
		\STATE
		 \textbf{goto} 1, using the improved $\phi$.
		
	\end{algorithmic}
\end{algorithm}

Along with classical regression models, we use deep neural networks for regression and classification tasks. In the next two sections we will provide a brief introduction to neural networks, in which we define neural networks and outline standard training procedures. We will then describe our implementation of such deep learning models (\ie choosing hyperparameters) for the computational problems at hand.  

\section{Deep learning models}\label{sec:neural_network}

In this section we provide a quick overview of neural networks for the mathematician who does not specialize in applied or computational mathematics. A good reference for this subject is~\cite{Goodfellow-et-al-2016}.

Section~\ref{sec:nn_approximates} establishes an analogy between neural networks and polynomial interpolation. Neural networks are defined in Section~\ref{sec:nn_define} together with the class of functions associated to them. In Section~\ref{sec:nn_descent}, we outline the method of gradient descent on this class of functions. 

\subsection{Neural networks to approximate functions}\label{sec:nn_approximates}

A neural network is essentially a framework for approximating functions. Let $\varphi \colon \Rr^n \to \Rr^m$ be an unknown ``target" function that we wish to approximate, and let $\Gamma_\varphi \subset \Rr^{n+m}$ be the graph of $\varphi$. Suppose that we have access to a finite subset $\ct$ sampled from $\Gamma_\varphi$. Assuming strong hypotheses on $\ct$ and $\varphi$, we can in some cases closely approximate or even recover $\varphi$. 

A classical example of such a scenario is univariate polynomial interpolation. If the target function $\varphi$ is known to be a polynomial with degree bounded above by some known $k$, then we can closely approximate $\varphi$ from $\ct$ (provided $\ct$ is sampled sufficiently well)
by solving a reasonably-sized linear system. This example demonstrates a good use-case for polynomial interpolation. However, the linear system could be defined by a Vandermonde matrix that is extremely ill-conditioned, making a solution very difficult and sensitive to noise.

Thus interpolation may not always be the most appropriate way to learn~$\varphi$. In many real-world scenarios (e.g. that encountered in this work), we cannot assume that $\varphi$ is polynomial. A function $\varphi$ that is \emph{not} sufficiently regular may require interpolating polynomials of very high degree. This can result in a high-dimensional linear system requiring a large matrix inversion, which can be a computationally prohibitive task. In this work, we relax our requirement of a perfect fit between model and data, and use a neural network architecture with a variation of gradient descent to solve such a regression problem.

In analysis, it is common practice to construct a class of functions $\cc$ which exhibits so-called good approximating properties. Once an appropriate class has been chosen, we try to find a sequence of functions $\{\varphi_n\}_{n=1}^{\infty} \subset \cc$ which converges in some sense to~$\varphi$. As will be discussed below, a chosen neural network architecture determines such a class\footnote{This class is by no means guaranteed to be unique or best for the task at hand.} $\cc$ and provides an algorithm for constructing an approximating sequence $\{\varphi_n\}_{n=1}^{\infty}$ using $\ct \subset \Gamma_\varphi$ that will hopefully converge to $\varphi$. This is more generally referred to in computer science as \emph{regression}.

\begin{remark}\label{rem:dont_approx_periods}
  We seek to approximate a complexity function (as in Section~\ref{sec:search_computation_graph}) associated to the period computations, rather than approximating the period function itself. This is because our intended use for the periods require hundreds of digits of accuracy and, ideally, rigorous bounds on error. Approximating functions with neural networks typically capture large scale features of a function. Attaining high precision, let alone bounding the error, is not one of the highest priorities of the method. Furthermore, we can better tolerate error in the complexity function as we will use it only as a heuristic to order our computations (Section~\ref{sec:informed_brute}). 
\end{remark}

\subsection{The class of functions associated to a neural network}\label{sec:nn_define}

We define a \emph{neural network} as a triplet $\cn=(\vc,E,A)$ where $\vc=(\vc_0,\dots,\vc_{k+1})$ is a sequence of real vector spaces with a fixed coordinate system $\vc_i = \Rr^{n_i}$, $E = (E_1,\dots,E_{k})$ is a sequence of non-linear endomorphisms $E_i \colon \vc_i \to \vc_i$ and $A=(A_0,\dots,A_{k})$ is a sequence of \emph{affine} transformations $A_i \colon \vc_{i} \to \vc_{i+1}$. 
By an affine transformation we mean the composition of a linear map $\vc_i \to \vc_{i+1}$ with a translation $\vc_{i+1} \to \vc_{i+1}$. 

\[
  \begin{tikzcd}
    \vc_0 \arrow[out=-60,in=-120,rrrr,"{ f_A }"]\arrow[r,"A_0","\text{linear}"'] & \vc_1 \arrow[out=120,in=60,loop,"E_1"] \arrow[r,"A_1","\text{affine}"'] & \dots \arrow[r,"A_{k-1}","\text{affine}"']& \vc_k \arrow[out=120,in=60,loop,"E_k"] \arrow[r,"A_k","\text{affine}"'] & \vc_{k+1} \arrow[out=120,in=60,loop,"E_{k+1}"] 
  \end{tikzcd}
\]

The non-linear endomorphisms $E_i$ are typically chosen to be of a very specific form in order to facilitate computations and to enable gradient propagation.
They do not all need to be the same function. Unless specified otherwise, we will always take each $E_i$ to be a \emph{rectified linear unit}, also denoted ReLU.

\begin{definition}
  For any $x \in \Rr$ the function $x^+ = \max(0,x)$ is called the \emph{rectifier}. For $W=\Rr^n$, \emph{the ReLU on $W$} is the non-linear map $W \to W$ defined by $(x_1,\dots,x_n) \mapsto (x_1^+, \dots , x_n^+)$.
\end{definition}

The \emph{architecture of a neural network} consists of the choice of $\vc$ and $E$. During the training of the neural networks, the architecture remains fixed and only the affine transformations $A$ are changed. The entries of the matrices representing $A_i$'s are called \emph{parameters} of the neural network $\cn$. 

When the architecture $(\vc,E)$ is fixed, we suppress it from notation and associate to each neural network $(\vc,E,A)$ the function $f_A \colon \vc_0 \to \vc_{k+1}$ defined as follows:
\begin{equation}
  f_A \colon v\mapsto  \underbrace{E_k \circ A_k}_\text{output layer}\circ E_{k-1} \circ \cdots \circ A_1 \circ  \underbrace{E_0 \circ A_0}_\text{input layer} (v)
\end{equation}
for some fixed $k$, referred to as a \emph{hyperparameter} of the network. The name hyperparameter refers to certain characteristics (e.g. the \emph{depth} $k$ of the network and the \emph{width} $\dim(\vc_i)$ of each layer) that define the class $\cc$ of functions allowed by the neural network.
We will call each $ A_i \circ E_i $ a \emph{layer} of the neural network $\cn$. We refer to layer $i=0$ (resp. $i=k$) as the \emph{input} (resp. \emph{output}) \emph{layers}; the remaining layers are called \emph{hidden layers}.
The naming suggests the layers' use and the asymmetry apparent in the construction of $f_A$. The affine transformations $A_i$ are parametrized by variables called \emph{weights} and variables called \emph{biases}.
In this setting we will equate a neural network simply with a composition of such layers.

It is remarkable that something as simple as the incorporation of ReLU functions vastly expands the space of functions that can be represented by neural networks. Without the non-linearity of the endomorphisms $E$, the function $f_A$ would simply be an affine transformation.

\subsection{Gradient descent using neural networks}\label{sec:nn_descent}

Fixing the architecture $(\vc,E)$ of a neural network yields a parametrized family of neural networks $(\vc,E,A)$, which can be associated to the family of functions $\cc = \{f_A \mid A\}$. 
For each $i>0$, the affine transformation $A_i$ has $n_in_{i+1}+n_{i+1}$ trainable parameters, so the total number of parameters in $f_A$ is
\begin{equation}
  N=\sum_{i=0}^{k} (n_i+1) n_{i+1}.
\end{equation}
That is, we have a parametrization $\Rr^N \tos \cc= \{f_A \mid A\}$. 
A~choice of transformation $A^{(0)} \in \Rr^N$ induces a neural network function ${f_{A^{(0)}} \colon \vc_0 \to \vc_{k+1}}$. 

Our immediate goal is to find a sequence $A^{(k)} \in \Rr^N$, $k \ge 0$, such that the sequence of functions $\{f_{A^{(k)}}\}_{k=0}^{\infty}$ converges in a sense to an approximation of our target function $\varphi \colon \vc_0 \to \vc_{k+1}$. We will first describe the distance measure that defines convergence for our experiments.

\subsubsection{Loss function}

Given a function $g \colon \vc_0 \to \vc_{k+1}$ we wish to quantify the goodness of our network's function approximation, \ie how far $g$ is from being equal to $\varphi$. The only information we are given about $\varphi$ is the finite subset $\ct \subset \Gamma_\varphi \subset \vc_0 \times \vc_{k+1}$ of its graph; this subset is called the network's \emph{training set}. As we have a fixed coordinate system on $\vc_{k+1}$, we will use the induced Euclidean norm $\norm{\cdot} \colon \vc_{k+1} \to \Rr_{\ge 0}$.

Let $P(\Gamma_\varphi)$ denote the set of finite subsets of $\Gamma_\varphi$. Let $\mathcal{L} \colon \Rr^N \times P(\Gamma_\varphi) \to \Rr_{\ge 0}$ be the \emph{loss function} defined as follows:
\begin{equation}
  \mathcal{L}(A;\ct) \colonequals \sum_{(t_1,t_2) \in \ct} \norm{f_A(t_1)-t_2}^2.
\end{equation}
As in the classical philosophy of regression, we say that \emph{to fit $\varphi$ well given $\ct$ is to minimize $\mathcal{L}$ with respect to $A$}. In the following we discuss two variations of a popular iterative method with the aim of achieving the minimization 
$$\argmin_{A\in \Rr^N} \mathcal{L}(A;\ct).$$

The following class of algorithms provides a sequence $\{f_{A^{(k)}}\}_{k=0}^{\infty}$ whose $\mathcal{L}$-values will hopefully (and in some limited cases, provably) be decreasing. 

\subsubsection{Gradient descent}

Given $\ct \subset \Gamma_\varphi$ we wish to find $A\in \Rr^N$ minimizing the error function $\mathcal{L} (\cdot,\ct)$. Our restriction to feedforward networks forbids feedback loops and, therefore, allows for an easy evaluation of the gradient $\nabla\mathcal{L} (\cdot,\ct)$ of $\mathcal{L} (\cdot,\ct)$ at any given point $A$ via backpropagation --- see \cite{Goodfellow-et-al-2016} for definitions and details.

Choose an initial point $A^{(0)} \in \Rr^N$ and a sequence of \emph{step sizes} $\gamma \colon \Nn \to \Rr_{>0}$. This sequence is typically either constant, or converging to zero. Inductively define the following sequence:
\begin{equation}
  A^{(k)} \colonequals A^{(k-1)} - \gamma(k) \nabla\mathcal{L} (A^{(k-1)};\ct), \quad k>0.
\end{equation}

In the applications we have in mind, the size of $\ct$ will be too large to make the execution of this method feasible. In examples where $\# \ct$ is large, one might opt for some variant of \emph{stochastic gradient descent}; one such method is described below.

\subsubsection{Stochastic minibatch gradient descent}

Stochastic minibatch gradient descent differs from gradient descent in that one trains on a random subset of $\ct$ at each step, instead of $\ct$ itself. To do this, fix a \emph{batch size} $b \in \Nn$, and define $A^{(k)}$ inductively as follows: 

Choose a random subset $\ct_k \subset \ct$ of size $b$ and let
\begin{equation}
  A^{(k)} \colonequals A^{(k-1)} - \gamma(k) \nabla\mathcal{L} (A^{(k-1)};\ct_k), \quad k>0.
\end{equation}

\subsubsection{Hyperparameter selection}\label{sec:hyperparameters_definition}

In order to use a neural network architecture $(\vc,E)$ to approximate the function $\varphi$, for which a subset $\ct \subset \Gamma_\varphi$ is known, one requires the following: An error function $\mathcal{L} $, step sizes $\gamma$, batch sizes $b$, distributions to randomly choose subsets of $\ct$ and to choose a starting point $A^{(0)}$. Collectively, these choices $(\vc,E,\mathcal{L} ,\gamma,b)$ are called \emph{hyperparameters}. This defines the class $\cc$ as described in Section~\ref{sec:nn_approximates}.

Selecting hyperparameters can be more of an art than a science. Although principled selection strategies have been proposed, e.g.~\cite{claesen2015hyperparameter} provides an excellent survey of some such methods, this problem remains largely unsolved for general learning problems. The rate of convergence of an iterative algorithm, and even whether the algorithm converges at all, can depend heavily on parameter choice. For example, a step size chosen too small will cause the algorithm to crawl slowly to a local minimum, whereas a step size that is too large might cause the algorithm to diverge.
Experimentation is required to select hyperparameters in a way so that the stochastic gradient converges rapidly and to a reasonable approximation of $\varphi$.
 
In the next section, we will discuss our implementation of such a learning algorithm, including how we chose the hyperparameters for the problem of approximating a complexity function as in Section~\ref{sec:search_computation_graph}.

\section{Implementation}\label{sec:implementation}

Now that we have reviewed the fundamentals of neural networks, we return to the problems explained in Section~\ref{sec:search_computation_graph}. 
We propose data-driven approaches to learning a proxy to the complexity function $\varphi$ of that section. 

The design space for our experiments includes a choice of labelled training dataset $\ct$, a statistical model (classical or neural), and hyperparameters that define that model.
Sections~\ref{sec:compscore}--\ref{sec:datagen} are concerned with the design of the dataset on which we will learn. 
Section~\ref{sec:features} compares useful intermediate representations of this dataset, a subdiscipline called \emph{feature extraction}. 
Section~\ref{sec:performance} discusses the experiments with $4$-and $5$-monomial quartic surfaces that motivated our choice of hyperparameters.

\subsection{Computability score} \label{sec:compscore}

The goal of the learning task in this section is to obtain a function $\phi\colon E \to [0,1]$ which assigns to each edge $e \in  E$ a probability that the computation $e \mapsto \cp_e$ of the period transition matrix will terminate on our hardware. 
For any such $\phi$ we will refer to $\phi(e)$ as a \emph{computability score of $e$}. 
The idea is to choose $\phi$ so that $\phi(e)$ is easy to evaluate but $\phi(e)$ is indicative of the computability of our entire algorithm ($e \mapsto \cp_e$)  on $e$.

\begin{remark}
One could, in principle, set $\phi=\frac{1}{1+\varphi}$ where $\varphi$ is the complexity function introduced in Section~\ref{sec:search_computation_graph}. However, the difficulty in explicitly computing $\varphi$ motivates us to find a more practical, data-driven solution.
\end{remark}

Assuming the kind of function $\phi$ we seek is moderately well behaved, we could approximate it via statistical learning methods using only finitely many pairs $(e,\phi(e))$. We will choose a random subset $E' \subset E$ and assign a value $\phi(e)$ for each $e \in E'$. The resulting set of pairs 
\begin{equation}\label{eq:training_data}
  \ct = \{(e,\phi(e) )\mid e \in E'\} 
\end{equation}
will be used to train a statistical model to obtain a modest guess for what $\phi$ should be. The remaining pairs $E\backslash  E'$ will be used for testing and validation.
We will next define the $ E$ and $\phi$ that we have chosen for our application to the problems introduced in Section~\ref{sec:problem_types}.

\subsection{Experimental input space $ E$}\label{sec:inputsp}
The abstract discussion in this chapter applies in the full generality of Section~\ref{sec:problem_types}.  Recall the notation of Section~\ref{sec:second_problem}, where $V_k \subset \Qq[x,y,z,w]_4$ denotes the \emph{$k$-nomial data set}, that is, the set of four-variable homogeneous polynomials that are the sum of $k$ distinct monomials all with coefficient $1$. 
For this section, we will constrain ourselves to fixed sets of polynomials --- \ie to the complete graphs on $V_4$ and $V_5$.

We denote by $ E$ the edges of the graph $V_k$ that is eventually to be traversed. For instance, the $4$-nomial data set is defined by the complete graph on $V_4$ and thus $\# E = {V_4 \choose 2}$.

\subsection{Experimental computability score $\phi$}\label{sec:datagen} 
In this section we define our choice of computability score $\phi$.
Experimentally we find that one critical subroutine of our larger algorithm $e\to\cp_e$ tends to present a computational bottleneck.
The global computation of $\cp_e$ requires computing as many ODEs as columns of $\cp_e$ and then numerically integrating them. We will let $\phi$ represent how long it takes to compute only the first of these ODEs on a fixed computer. That is, $\phi\colon E\to \mathbb{R}_{\geq 0}$ is defined so that $\varepsilon(e)$ is the computation time for the first ODE required for $\cp_e$.

\begin{figure}
\includegraphics[width=\linewidth]{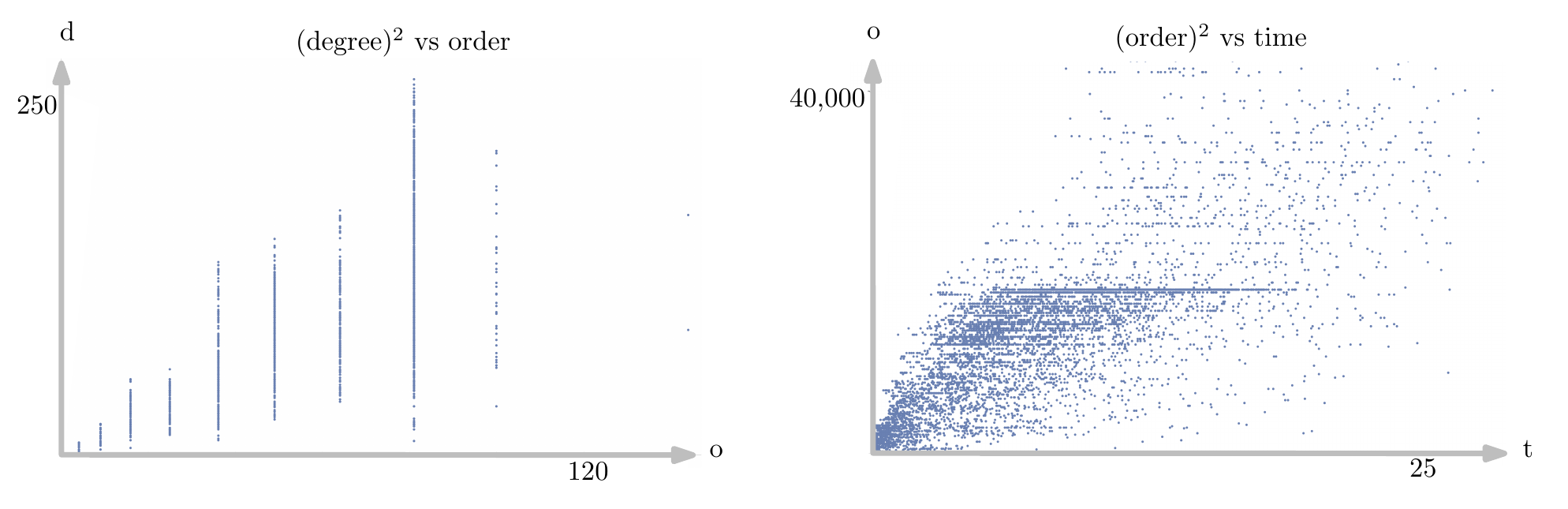}
\caption{$5$-nomial data. Notice that the squares of degree and order are plotted. Shows a very loosely quadratic relationship between time and order, and between order and degree. This correlation motivated our sole use of time as a label.}
\label{fig:odt}
\end{figure}

\begin{remark}
We experimented with including additional label information in training, such as the degree and order of the first ODE.
However, we did not observe an advantage to using this additional data in training our neural networks. 
This may be attributed to the visible relation between time, degree and order in Figure~\ref{fig:odt}.
\end{remark}

\subsubsection{A practical complexity measure}\label{sec:fragment} 

We generate many training samples of $\phi$, evaluated on edges of the graphs based on $V_4$ and $V_5$.
Since $\phi(e)$ can be arbitrarily large, we choose a threshold and terminate the computation of $\phi(e)$ after $30$ seconds.
In that case, we know $\phi(e) \ge 30$.
To capture this process we define a binary label $\beta(e) \in \{0,1\}$, where $\beta(e)=0 $ if $\phi(e) <30$ (\emph{successful computation}) and $\beta(e)=1$ if $\phi(e) \ge 30$ (\emph{failed computation}).

We will refer to the resulting edge-time correspondences $\left\{\left(e,\beta(e)\right)\right\}$ for $V_k$ as \emph{$k$-nomial data sets}.
These pairs will either be used for training or testing, in which case we may refer to $e$ as an input data sample,
and to $\beta(e)$ as that sample's label. Our architecture choice is in part guided by the structure of this data. In our learning task, it ends up being easier to learn the binary classification $\beta$ rather than the nuanced function $\phi$ itself; in fact, in learning we obtain a proxy for $\phi$ given by the per-class probability distribution on $E$. Thus we replace our regression problem (approximate $\phi$) with a binary classification problem (approximate $\beta$, and get a proxy for $\phi$ en route). We refer to these two functions interchangeably in the development of our strategy.

We must be careful with how we represent an edge $(f,g)$ (i.e, a pair of quartic surfaces). Finding a good representation is called \emph{feature extraction} and can be seen as enlarging the input space $\vc_0$ in a meaningful way. 
As the following section details, we are able to compute extra data associated to $(f,g)$ that empirically relates to the complexity $\varphi(f,g)$. 

\subsection{Feature extraction}\label{sec:features}

We approximate the computability score function $\phi \colon  E \to [0,1]$ via a statistical model that takes as input a vector representation of $e \in  E$. We represent each edge $e$ by the concatenation of the coefficient vectors of the two polynomials that are the endpoints of that edge. 
It must be noted that in using this vector representation, we have discarded some information: the value of $\phi$ is linked inextricably to the fact that endpoints of $e$ are polynomials, a characteristic of which the model is no longer aware. This guides the idea that our chosen learning method can better guess the value of $\phi(e)$ if it is provided more than just the coordinate representation of $e$. 

One additional piece of domain information that we found useful was the \emph{first-order} Gauss--Manin connection, see Section~\ref{sec:first_order_gm}. We can efficiently compute the first-order Gauss--Manin connection at a few points of the pencil corresponding to $e$. In practice, the complexity of these matrices correlates with the computability score of $e$, as shown in Sections~\ref{sec:complexity_function} and~\ref{sec:learning_computability}. 

We will use the following notation for these matrices. First, fix $t_1,\dots,t_s \in \Cc$. For each $e$ let $M_e = (M_{e,1},\dots,M_{e,s})$ be a sequence of such matrices, where $M_{e,i}$ is the first order Gauss--Manin connection evaluated at $t=t_i$ on the pencil corresponding to $e$. 
We will design our models to work with $(e,M_e)$. In practice, we take $s=1$ or $s=2$ with $t_1=0$ and $t_2=1$ because we observed little benefit in increasing $s$ further.

\subsection{Studying the complexity of the input}\label{sec:analyzing_matrices}

When the polynomials used in $E$ have rational coefficients, the connection matrices $M_e$ will also be rational-valued. 
For example, when we are dealing with quartic surfaces defined over $\Qq$, we have that $(e,M_e)$ is an element of $\Qq^{n}$,
where $n=2\times 35+ s \times 21\times 21$. 
In this section, we will consider various measures of complexity of the matrix $M_e$. 
Any correlation between ``easy'' statistics (defined in the next section) of $M_e$, and the algorithm-fragment runtime $\phi(e)$
is not obvious. This motivates the use of more complex data-driven models such as neural networks. 

\subsubsection{Complexity of cohomology matrices}\label{sec:complexity_function}

Because the standard implementation of neural networks work with floating point arithmetic, the subtleties of computing with a rational number are lost. As a large portion of the computation $e \mapsto \cp_e$ is exact, the computation time is affected by the ``height'' of the rational numbers involved. For this reason, we will modify the entries of $M_e$ to better represent the complexity of its entries.

Let us define the following function on rational numbers:
\begin{equation}\label{eq:psi}
  \begin{aligned}
    \psi \colon \Qq \to& \Rr_{\ge 0} \\
    \frac{m_1}{m_2} \mapsto& \log(\abs{m_1}) + \log(m_2)
  \end{aligned}
\end{equation}
where $m_1,m_2 \in \Zz$, $m_2 > 0$ and $\lcm(m_1,m_2) = 1$.
The value $\psi(m)$ of a rational number $m$ is a more faithful representation of the complexity of computing with $m$ then would be a floating point approximation of $m$. The following variation will also be used:
\begin{equation}
  \begin{aligned}
    \psi_{\text{entropy}}\colon \frac{m_1}{m_2} \mapsto& \log(\abs{m_1})^2 + \log(m_2)^2. \\
  \end{aligned}
\end{equation}

Various complexity statistics can be extracted from a complexity matrix~$M$. Each statistic is a function $\Psi(M)$ on a $3$-tensor $M=(M_{ijk})$ of rational numbers, such as $M_e$. We list three options on Table~\ref{tab:statdefs}. 

\begin{table}[h]
\begin{tabular}{ |l|rl| }
\hline
Measure & Definition&\\
\hline
Sum &$ \Psi_s(M):=$&$ \sum_{i,j,k} \psi(M_{ijk}) $  \\ 
 Entropy & $ \Psi_e(M):=$&$ -\sum_{i j k} \psi_{\text{entropy}}(M_{i j k})  $ \\
 Length Nonzero & $ \Psi_l(M):=$&$ \text{len}(m\in \psi M: m\neq 0)$  \\ 
 \hline
\end{tabular}
\caption{Three of the examined complexity measures.}
\label{tab:statdefs}
\end{table}

\begin{figure}
      \centering
\includegraphics[width=\linewidth]{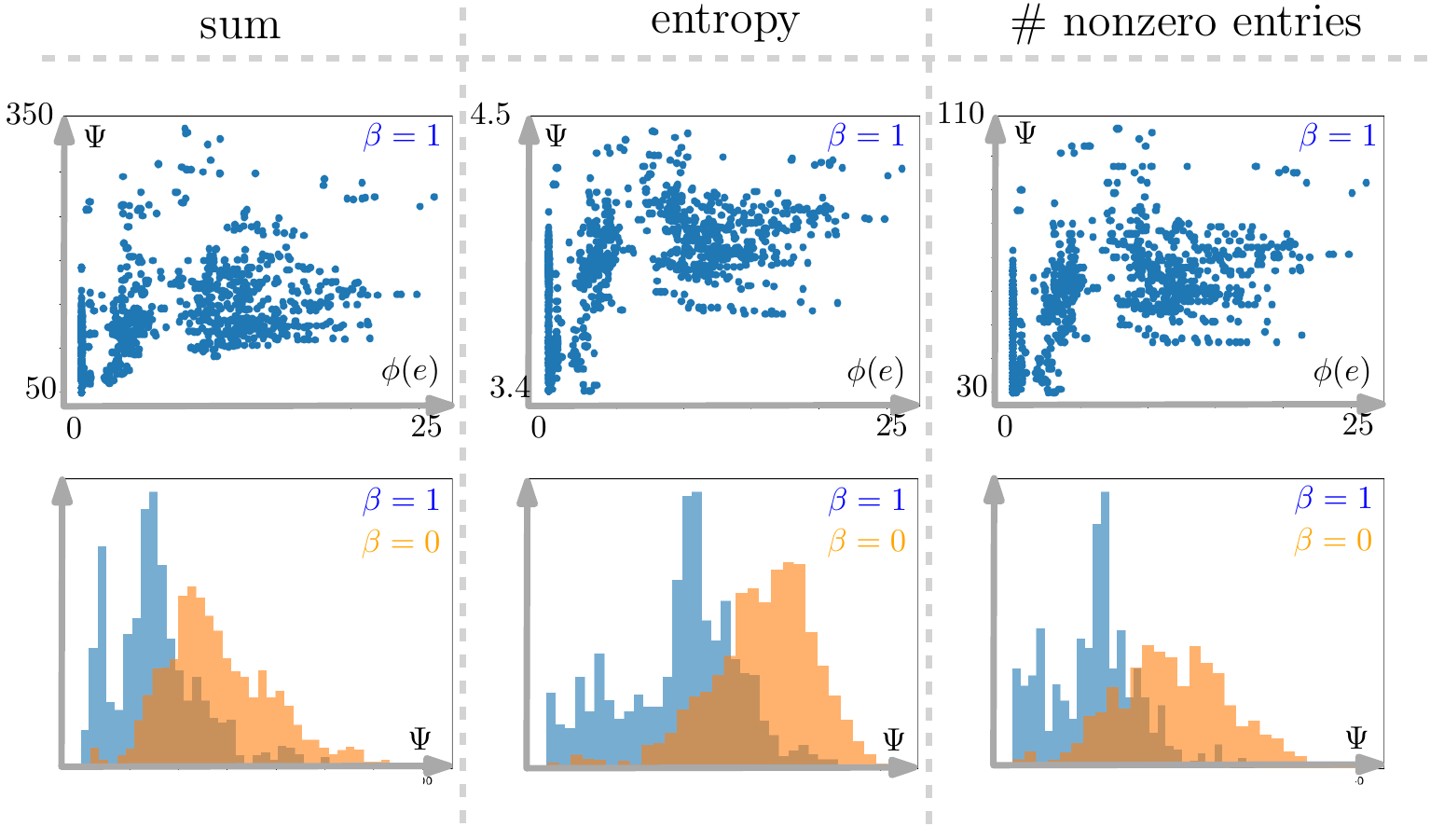}
\caption{\textbf{Complexity measure versus time.} Dependence between the matrix statistics and computation time on the 4-nomial dataset. \emph{Top row:} $\Psi(M_e)$ versus $\phi(e)$ on successful edges. 
\emph{Bottom row:} Histograms of matrix statistics $\Psi(M_e)$, with the successful-edge distribution ($\beta(e)=1$) denoted in blue, and the failing-edge distribution ($\beta(e)=0$) denoted in orange.}
       \label{fig:cohomstats}
\end{figure}

Each column of Figure~\ref{fig:cohomstats} corresponds to an entry in Table~\ref{tab:statdefs}. For each $\Psi$, we plot $\Psi(M_e)$ against the time it takes to compute the algorithm fragment defining $\phi$, for those edges such that $\beta(e)=1$. The edges for which $\beta(e)=0$ are omitted from the top row of Figure~\ref{fig:cohomstats} because we had to terminate their computation prematurely, effectively assigning them all the same value of 30. The bottom row of Figure~\ref{fig:cohomstats} shows that successful edges do tend to have lower matrix statistics than failing edges. However, these distributions are not bimodal enough to make the statistics good classifiers in isolation. 

On the other hand, we see no striking patterns between $\phi(e)$ and $\Psi(M_e)$ in the first row of Figure~\ref{fig:cohomstats}. From this we conclude that the matrices $M_e$ are useful in terms of classifying successes from failures (\ie approximating $\beta$), but their statistics alone are not sufficient to regress (\ie approximate $\phi$) within the class of successes.

\begin{figure}
      \centering
      \includegraphics[width=\linewidth]{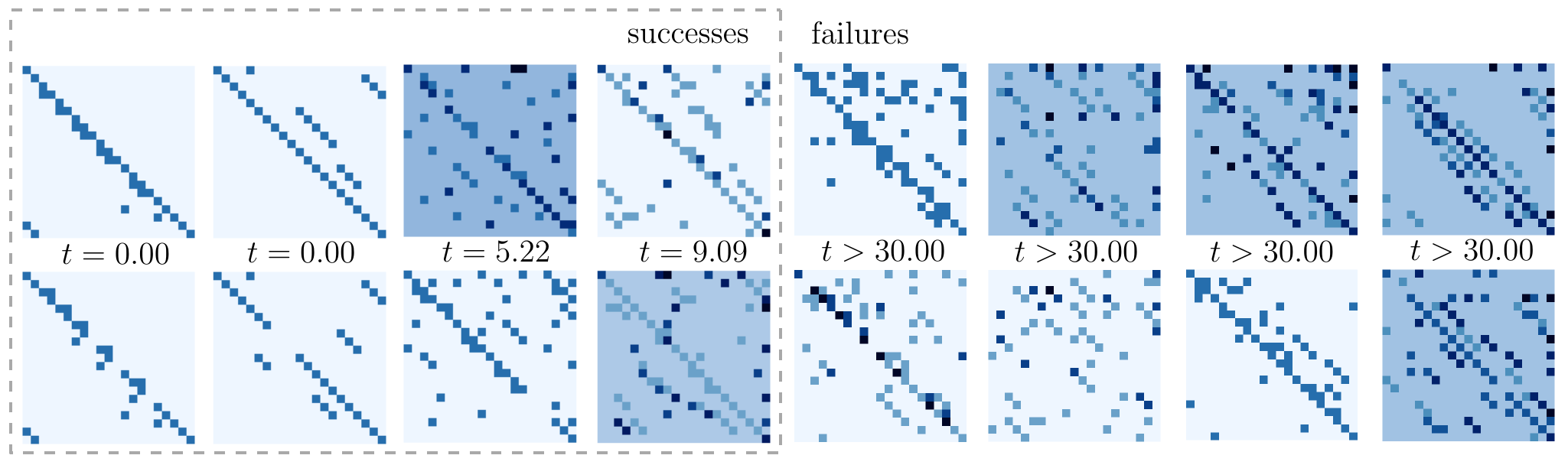}
      \caption{How our convolutional neural networks see $M_e$'s. Top row represents $\psi(M_{e,1})$ and the bottom represents $\psi(M_{e,2})$, with $\psi$ as in~\eqref{eq:psi} evaluated entrywise. Darker shades indicate larger values of~$\psi$. 
    }
       \label{fig:cohommats}
\end{figure}

We can gain more insight by viewing each tensor $M_e$ as a multi-channel image. 
We can visualize a matrix $(M_{ij})$ as a rectangular image with the $ij$-th entry colored a shade of blue:
the darker the shading, the larger the value of $\psi(M_{ij})$.
See Figure~\ref{fig:cohommats}, which has $M_{e,1}$ in the first row and $M_{e,2}$ in the second row for four $e$ from the $4$-monomial data set.
We will use convolutional neural networks to better convey the spatial relationship of entries in a matrix to the neural network. 

\subsection{Preprocessing the data}
\subsubsection{Dimension reduction: principal component analysis}\label{sec:pca}

In the edge sets $E$ we consider, the pair of polynomials in $e$ for each $e \in  E$ are sparse.
As a result, we are using an unnecessarily large ambient space to represent $e$. 
This in turn, dilutes the capacity of a neural network to learn.
Instead we perform a \emph{principal component analysis} (PCA) on a given $ E$ to compress the data stored in $e$ by projecting onto a subspace with minimal loss in information.

\begin{figure}
\includegraphics[width=3in]{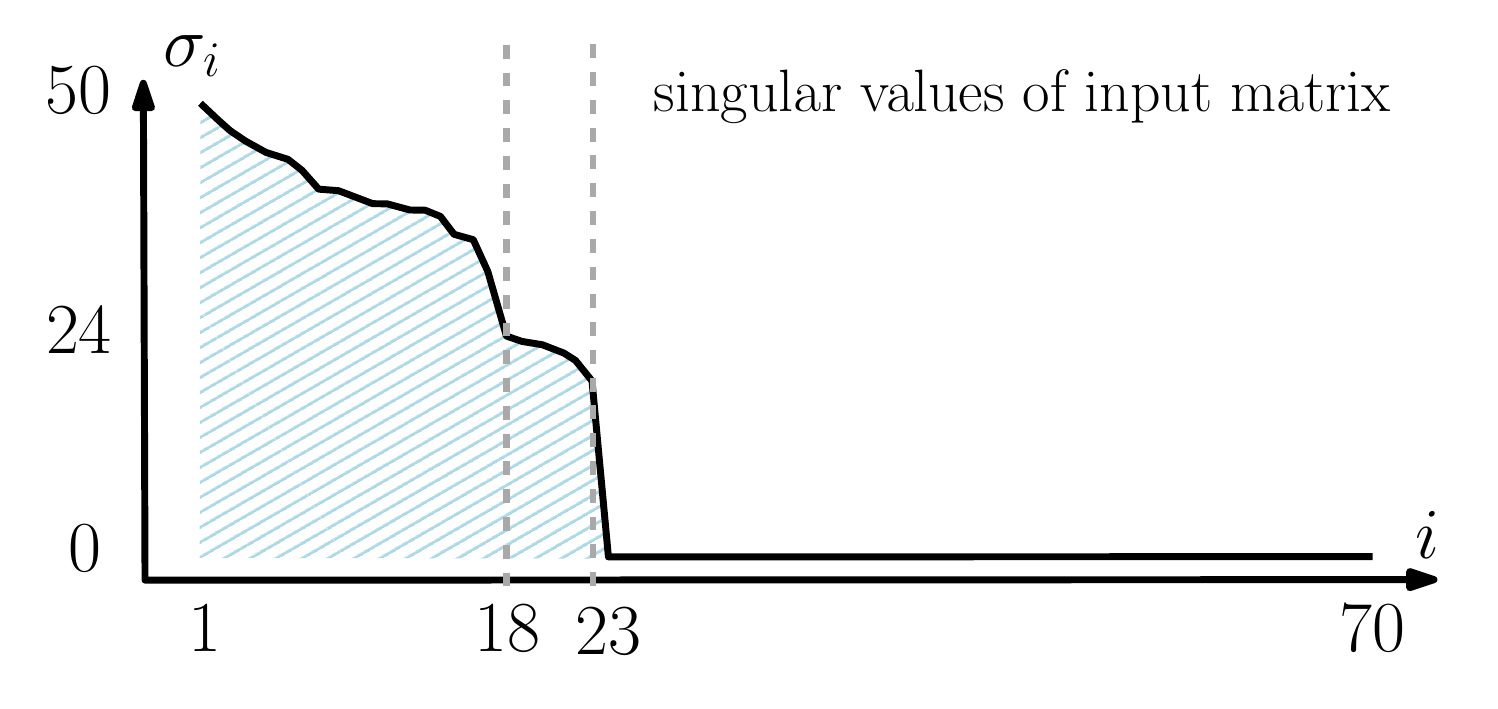}
\caption{The distribution of the singular values of the complete 4-nomial graph.}
\label{fig:pcs}
\end{figure}

When considering quartic surfaces, $E$ lies in a $70=35+35$ dimensional ambient space. However, projecting onto a well chosen smaller subspace cuts down the dimension with minimal loss of information. For example, Figure~\ref{fig:pcs} shows that the first $23$ principal components almost entirely recovers the coordinates of edges in the complete graph over $V_4$. This number $23$ works well also on $V_5$. We thus compress each $e$ to a $23$ dimensional vector, down from $70$.

\subsubsection{Balancing the dataset}

Unfortunately, even the algorithm fragment of Section~\ref{sec:fragment} fails to terminate in the vast majority of pairs we considered. As a result, a training set $\ct$ as in~\eqref{eq:training_data} will consist almost entirely of failed edges of the form $(e,0)$. This incentivizes the neural network to produce the constant $0$ function as an approximation of $\phi$. 

As a remedy, we employ a standard method to disincentivize the neural network from converging to a constant. This method is done by over-sampling the under-represented class (here, edges $e$ with $\beta(e)=1$) so that the training set consists of an equal portion of both classes.

\subsection{Learning the computability score}\label{sec:learning_computability}

A crucial decision is whether to use neural networks at all. Classical statistical methods are simpler to operate and when they are more successful, they can give additional insight about the dataset. However, neural networks have a broader applicability and they can succeed when the classical methods fail.  In Section~\ref{sec:classical_methods} we show that we are in the latter case, the neural networks consistently outperform the classical methods.

In this subsection we will work with the $4$-monomial dataset, i.e. the complete graph on $V_4$. We tried several classification techniques as candidates for ``learning'' $\phi \colon  E \to[0,1]$. It turns out that we get a good estimate for $\phi$ if we just try to approximate the classification function $\beta \colon E \to \{0,1\}$. The methods we consider output functions $f\colon E \to [0,1]$ and we must couple them with a cut-off value $\tau$ so that $\beta(e) = 0$ is most likely correlated with $f(e) \le \tau$. 

\begin{figure}
\centering
\includegraphics[width=\linewidth]{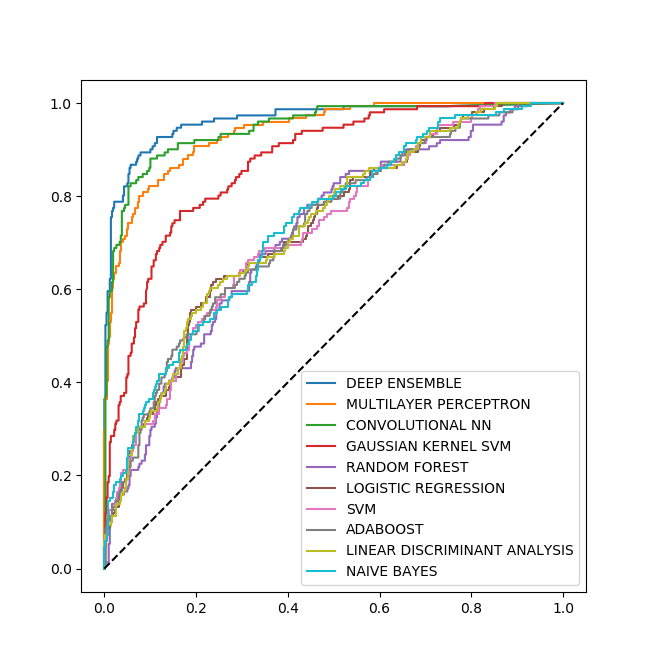}
\caption{\textbf{True positive rate versus true negative rate.} Receiver operating characteristic curves of some binary classifiers, on 4-nomials. The top performers are deep learning methods.}
\label{fig:roccurves}
\end{figure}

In order to rank different methods, we drew in Figure~\ref{fig:roccurves} the \emph{receiver operating characteristic curves} of ten different methods. The curve for each method is obtained by varying the cutoff value $\tau$, with true positive rate and true negative rate as the axes. The closer a curve is to the top left vertex, the better the corresponding method performs. The dotted line is the idealized curve for the method of random guessing. 

The best-performing method (deep ensemble) is a composition of the next best two neural network strategies, namely multilayer perceptron and convolutional NN. The fourth place is occupied by a method that is related to both neural networks and classical statistical methods (Gaussian kernel SVM), this is reflected here by its performance. The remaining six are classical methods, all of which underperform in this task. In Section~\ref{sec:classical_methods} we describe these methods in greater detail.

\subsubsection{Classical statistical methods}\label{sec:classical_methods}

We tried seven standard binary classifiers to predict $\phi$ from the 4-nomial training set $\ct$. Since they were clearly outperformed by the neural networks, we will not discuss them in depth.  The methods we used are:
logistic regression with an $L_2$ penalty,
a regularized support vector classifier with a linear kernel and regularization parameter~$1$,
a regularized support vector classifier with degree-2 radial basis function kernel and regularization parameter~$1$,
a random forest with 10 trees, per-tree maximum depth 5,
an AdaBoost classifier with at most 50 estimators,
linear discriminant analysis,
and Gaussian Naive Bayes.

Figure \ref{fig:roccurves} suggests that the deep classifiers will outperform the classical methods. This may simply be caused by the fact that deep learning methods have more internal parameters and can, therefore, provide better approximations.

\subsubsection{Deep neural networks}\label{sec:deep}

We will introduce two neural networks and then combine their results into an ensemble method. The first one, a neural network, follows the standard formulation in Section~\ref{sec:neural_network}; which is a multi-layer perceptron (MLP). In dealing with the quartic fewnomial dataset, we decided on an architecture with five hidden layers, each of width $100$. As input, it takes only the edges $e \in  E$ after compression via principal component analysis as in Section~\ref{sec:pca}. Its output is a single value in $\Rr$. To restrict the codomain to the interval $[0,1]$ we apply the inverse logit function. After training, the neural network gives an approximation $\phi_{\text{MLP}} \colon  E \to [0,1]$ of the computability score. 

\begin{figure}
\centering
\includegraphics[width=\linewidth]{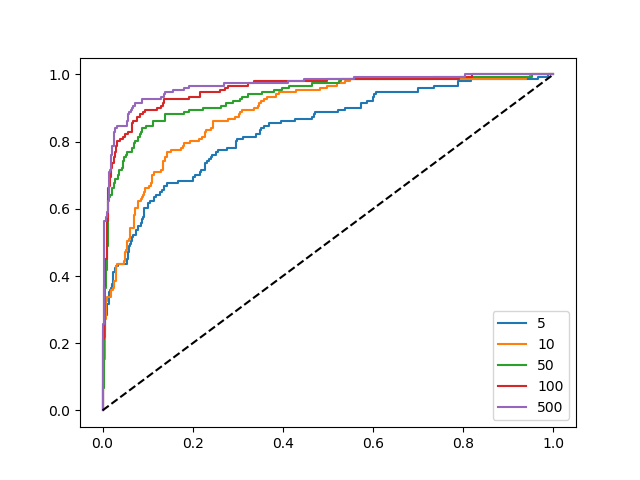}
\caption{Receiver operating characteristic curves of the multilayer perceptron (MLP) on the 4-nomial data. The key denotes the widths for the first three layers of each MLP.}
\label{fig:rocmlp}
\end{figure}

Figure~\ref{fig:rocmlp} shows the consequence of changing the dimension of the first three hidden layers. Other parameters have also been chosen by considering such figures to optimize predictive power against performance.

Our second neural network is a two-channel convolutional neural network, a variation of the standard neural network explained in Section~\ref{sec:neural_network} to better detect patterns in visual data. This neural network will be trained using the $3$-tensors $M_e$ encoding the first order Gauss--Manin connections. We will however apply the complexity function $\psi$ from Section~\ref{sec:complexity_function} to each entry of $M_e$ before giving it as input. The output is adjusted as with the first neural network (MLP) above so that, after training, we obtain an approximation $\phi_{\text{CNN}} \colon  E \to [0,1]$. 
More details about both network architectures are available in the supplementary code\footnote{See \url{https://github.com/a-kulkarn/period_graph}}.

\begin{figure}
\includegraphics[width=4in]{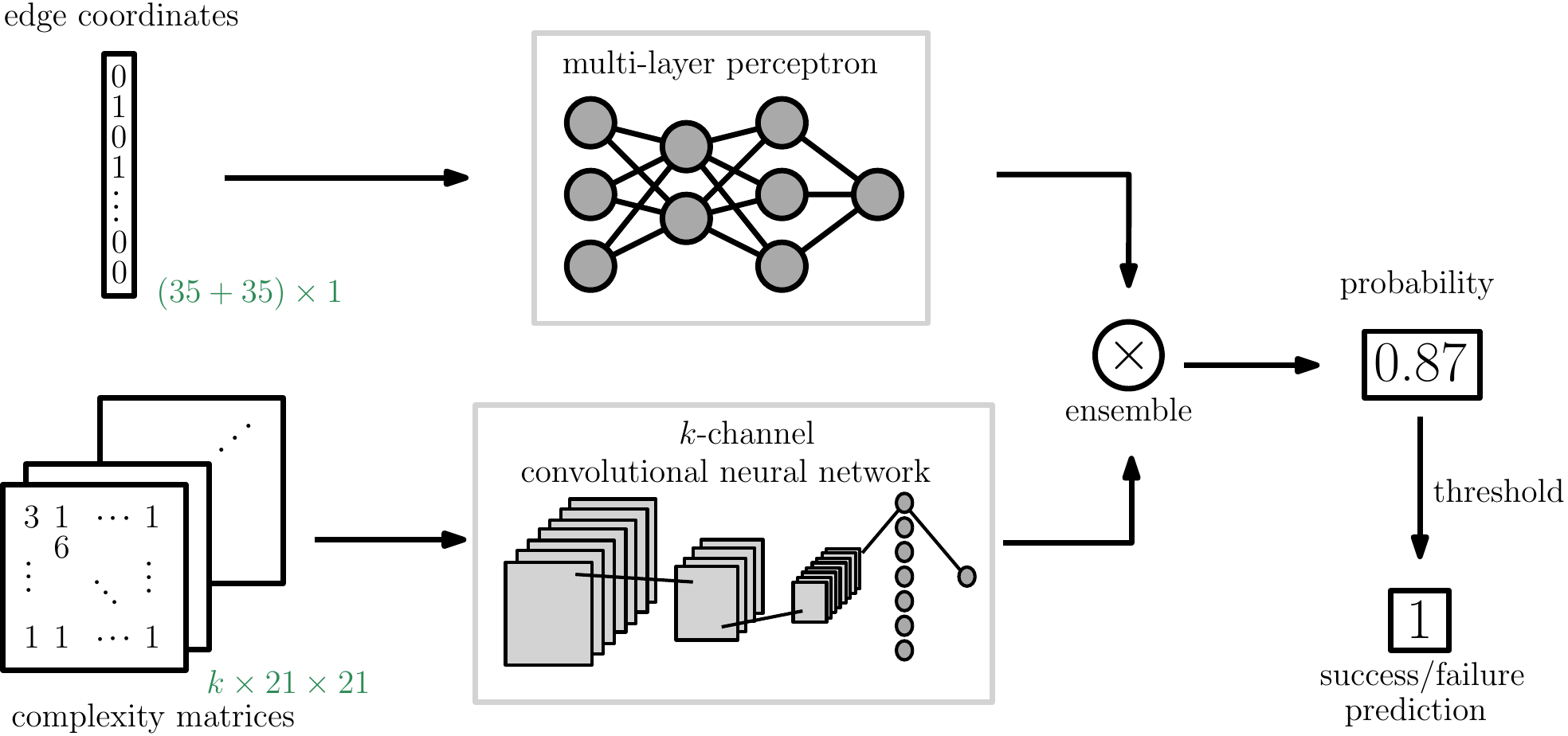}
\caption{Schematic of the deep ensemble binary classifier.}
\label{fig:netdiag}
\end{figure}

We improve on the approximations of the two neural networks by defining the function 
\begin{equation}\label{ensemb}
  \phi_{\text{ensemble}}:=\phi_{MLP}\cdot\phi_{CNN}.
\end{equation}
This is what we call the ``deep ensemble'' method (or just ensemble method), as illustrated in Figure~\ref{fig:netdiag}.  Being the product of two functions, which are essentially probability functions, the function $\phi_{\text{ensemble}}$ is more cautious in returning a value close to $1$. We chose this approach because attempting the computation $e \mapsto \cp_e$ for edges that do not terminate can be very costly. We prefer a computability score that has a low false positive rate.

\subsection{Performance on datasets}\label{sec:performance}

We now describe our experimental setup and the performance of our neural networks. Recall that $V_n$ is the set of smooth quartics that are the sum of $n$ monomials. 

\subsubsection{Four monomial quartics}\label{sec:4nom_performance}

Consider the edges $E$ of the complete graph with vertices $V_4$. Note $\# V_4 = 108$ and $\# E = {108 \choose 2} = 5778$. These numbers are small enough that we could run our test computation (as in Section~\ref{sec:fragment}) on all edges in $E$. This gives us complete information and allows us to evaluate the performance of our neural network.

For $\alpha \in (0,1)\subset \Rr$ take a random subset $E'\subset E$ for which $\# E' \sim \alpha \cdot \# E$. We train the ensemble neural network defined in~\eqref{ensemb} on $E'$ and test it on $E'' \colonequals E \setminus E'$. 
We wish to know whether the neural network correctly predicts if the edges in $E''$ can be successfully integrated in a short amount of time. 
We list in Table~\ref{fig:alpha} the percentage of \emph{true negatives (TN), false positives (FP), false negatives (FN) and true negatives (TN)}, for various values of $\alpha$. 
The neural networks learns to predict the answers with high level of accuracy even with training on a small fraction of the data.

\begin{table}[h!]
  \centering
      \begin{tabular}{cccccc}
      $\alpha$ & TN  (\%) & FP  (\%) & FN   (\%) & TP  (\%) & (TP+TN)/(FP+FN) \\ \hline
0.3            & 61.78    & 17.53    & 2.51      & 18.18    & 3.99 \\
0.5            & 57.65    & 14.60    & 3.86      & 23.89    & 4.42 \\
0.7            & 50.94    & 16.19    & 3.19      & 29.68    & 4.16
      \end{tabular}
  \caption{The performance of the ensemble neural network on $V_4$ with varying proportion $\alpha$ of edges used for training.}
  \label{fig:alpha}
\end{table}

Taking $\alpha=0.9$, in Figure~\ref{fig:rocmlp} we plotted the ROC curves of the multilayer perceptron component of the ensemble for different widths of its first three layers. We decided on a width of $500$ based on this figure.

\subsubsection{Four monomial quartics}\label{sec:5nom_performance}

The performance of the ensemble method trained solely on $V_4$ did not perform well on $V_5$.
One plausible reason is that the first order Gauss--Manin connections on $V_4$ are all quite simple, whereas on $V_5$ there is a broader range of complexity displayed by these matrices.
In other words, the structure in $V_4$ does not extrapolate well to $V_5$.

To address this, we retrained the ensemble method on $5$-monomials. Consider the edges $E$ of the complete graph with vertices $V_5$. This is a much larger set, since $3348 = \# V_5 \gg \# V_4 = 108$, so $\# E = 5,602,878$. With the ratio $\alpha$ defined as in Section~\ref{sec:4nom_performance}, we display the performance for various $\alpha$ in Table~\ref{fig:alpha2}. 

\begin{table}[h!]
  \centering
      \begin{tabular}{cccccc}
      $\alpha$ & TN  (\%) & FP  (\%) & FN   (\%) & TP  (\%) & (TP+TN)/(FP+FN) \\ \hline
0.3            & 83.06  & 1.80   & 0.69    & 14.44   & 39.05 \\
0.5            & 84.48  & 1.63   & 0.49    & 13.40   & 46.19 \\
0.7            & 83.70  & 1.52   & 0.46    & 14.32   & 49.48 
      \end{tabular}
  \caption{The performance of the ensemble neural network on $V_5$.}
  \label{fig:alpha2}
\end{table}

\section{Application}\label{sec:applications}

In this section we will give an application of our software package and analyze the performance improvement of using neural networks.

\subsection{Five monomials}

One application of our software is to the set $V_5$ of smooth polynomials, each of which are the sum of five monomial terms, with coefficients $1$, see Section~\ref{sec:second_problem} for this notation. Our goal is to compute the Picard numbers of these $5$-nomial quartics and to find unexpected isomorphisms in this set.

\subsection{List of results}

Note that the results of this section depend on finding integral linear relations between periods that are only known approximately. Therefore, the results below may contain errors due to insufficient precision. We tried to mitigate this risk by working with $300$~digits of precision, when we could attain it. 

Of $3348$ smooth $5$-nomial quartics, there are only $161$ $S_4$-symmetry classes. Out of this $161$ we could reach $154$. We computed their Picard numbers and we list their frequency below.

\begin{table}[h!]
  \caption{Picard number frequency for $5$-nomial quartics}
  \label{tab:pranks}
  \begin{center}
    \resizebox{\textwidth}{!}{%
    \begin{tabular}{r|cccccccccccccccccccc}
     $\rho$      & 1  & 2 & 3 & 4   & 5 & 6   & 7  & 8   & 9 & 10  & 11 & 12  & 13 & 14  & 15  & 16  & 17  & 18  & 19  & 20 \\ \hline
 frequency & 8 & 6 & 0 & 32 & 1 & 9 & 2 & 13 & 0 & 22 & 0 & 9 & 1 & 13 & 4 & 7 & 1 & 20 & 6 & 0 
    \end{tabular}}
  \end{center}
\end{table} 

  In contrast to the $180,000+$ quartics in the database given in~\cite{lairez-sertoz}, we are essentially after $161$ quartics. However, the database in \emph{loc.\ cit.}\ was built by random exploration to find easy to compute quartics. Here, our quartics $V_5$ are fixed. For this reason, we were forced to use an order of magnitude more computation time to build our graph (a CPU year versus decade). In fact, some of these quartics are so difficult to reach that our ``optimal'' tree $T$ spanning $154$ $S_4$-classes has a diameter of length $21$.

  \begin{remark}\label{rem:crystal}
  The Picard number for every $5$-nomial quartic in this database was verified using a complementary method that uses the crystalline cohomology of finite reductions~\cite{costa-sertoz}. This method produces guaranteed upper bounds. For each of our polynomials, we ran through as many prime reductions as was necessary to have the minimum of the upper bounds attained thrice. In each case, this minimum agreed with the Picard numbers we computed.
  \end{remark}

  \begin{remark}
    As we compute the Picard lattice and not just the Picard number, we can use~\cite[\S 3]{lairez-sertoz} to find classes of smooth rational curves in the Picard group. We observed that only $7$ of the $154$ quartics had Picard groups that could not be generated over $\Qq$ with the polarization and classes of smooth rational curves of degree $\le 3$. 
  \end{remark}

\subsubsection{The missing vertices}

While our search method succeeded for the vast majority of quartics,
we were unable to reach seven of the 5-nomial quartics in $V_5$, up to isomorphism.
These seven quartics are listed in Table~\ref{tab:missing}. 
Moreover, we list the Picard number bound obtained by {\tt crystalline\_obstruction}~\cite{costa-sertoz} as explained in Remark~\ref{rem:crystal}. 

We also tried to brute force every edge from $V_4 \cup V_5$ to establish a connection to the three quartics in this table with Picard number~$2$. Allowing three hours per connection to find only the first ODE succeeded only in connecting these vertices to one another. Naturally we could not try every edge with the three hour time limit. Therefore, it is conceivable but very unlikely that good connections exist.

This prompts us to ask: \emph{Which feature of these quartics is responsible for making them inaccessible?}

\begin{table}[h!]
  \caption{Picard numbers for the $7$ missing quartics}
  \label{tab:missing}
  \begin{center}
    \begin{tabular}{MM}
      \text{Polynomial}                & \text{Picard number} \\
    x^3y + y^3z + y^3w + z^3w + xw^3   & \le 2 \\
    xy^3 + z^4 + x^3w + y^2zw + xw^3   & \le 2 \\
    x^4 + y^3z + xz^3 + x^3w + yw^3    & \le 2 \\
    y^3z + xyz^2 + xz^3 + x^3w + yw^3  & \le 3 \\
    x^3y + y^3z + z^3w + z^2w^2 + xw^3 & \le 3 \\
    x^2y^2 + x^3z + yz^3 + y^3w + xw^3 & \le 18 \\
    xy^3 + x^3z + xyzw + z^3w + yw^3   & \le 19
    \end{tabular}
  \end{center}
\end{table} 

\subsubsection{Isomorphism classes}


Using the Torelli theorem for K3 surfaces~\cite{looijenga-peters} we can check if the K3 surfaces in our list admit non-trivial isomorphisms. Here, we work only with $154$ (of $161$) representatives of the $S_4$-symmetry classes for which we could compute the periods. We compared their period vectors modulo an integral change of basis for homology. The method of computation is described in~Section~\ref{sec:method_to_compute_iso_classes}. We found $9$ isomorphism classes of size $2$ and $2$ isomorphism classes of size $4$, all other isomorphism classes appear to be of size $1$. We display the non-trivial isomorphism classes in Table~\ref{tab:iso_classes}. 

\begin{table}[h!]
  \caption{The $11$ non-trivial isomorphism classes}
  \label{tab:iso_classes} 
  \begin{center}
    \resizebox{\textwidth}{!}{%
    \begin{tabular}{RL}
y^3z + yz^3 + x^3w + xw^3 + w^4 & y^4 + z^4 + x^3w + xw^3 + w^4 \\
\hline
y^4 + y^2z^2 + z^4 + x^3w + w^4 & y^3z + y^2z^2 + yz^3 + x^3w + w^4 \\
\hline
y^4 + z^4 + x^3w + yzw^2 + w^4 & y^3z + yz^3 + x^3w + yzw^2 + w^4 \\
\hline
y^4 + z^4 + x^3w + xz^2w + xw^3 & x^4 + y^4 + z^4 + yzw^2 + w^4 \\
\hline
y^4 + z^4 + x^3w + xyzw + w^4 & y^3z + yz^3 + x^3w + xyzw + w^4 \\
\hline
x^4 + y^4 + z^4 + zw^3 + w^4 & y^4 + yz^3 + z^4 + x^3w + xw^3 \\
\hline
y^4 + x^2yz + z^4 + x^3w + w^4 & y^4 + yz^3 + x^3w + xz^2w + xw^3 \\
\hline
y^3z + y^2z^2 + z^4 + x^3w + xw^3 & y^4 + z^4 + x^3w + x^2w^2 + w^4 \\
\hline
y^3z + yz^3 + x^3w + yzw^2 + xw^3 & y^4 + z^4 + x^3w + yzw^2 + xw^3 \\
\hline
y^3z + y^2z^2 + yz^3 + x^3w + xw^3 & y^4 + y^2z^2 + z^4 + x^3w + xw^3 \\
y^4 + z^4 + x^3w + x^2w^2 + xw^3 & x^4 + y^4 + z^4 + z^2w^2 + w^4 \\
\hline
x^4 + y^4 + z^4 + xyzw + w^4 & y^4 + z^4 + x^3w + xyzw + xw^3 \\
y^3z + yz^3 + x^3w + xyzw + xw^3 & y^4 + xz^3 + x^3w + xyzw + zw^3
    \end{tabular}}
  \end{center}
\end{table}



\subsubsection{Endomorphism fields}

We also computed the endomorphisms of the transcendental lattice of a K3 from its periods. We use the argument in~Section~\ref{sec:method_to_compute_iso_classes} with $X_1=X_2$ in order to compute these. The endomorphism ring $E$ of the transcendental lattice of a K3 is always a field, either of real or complex multiplication~\cite{zarhin--k3}. We did not observe any real multiplication surfaces. These are notoriously hard to find~\cite{EJ-RM}.  In Table~\ref{tab:endo} we list the polynomials $f(t)$ for which $E\simeq \Qq[t]/f(t)$ and the number of times this endomorphism field was realized among our $154$ $S_4$-symmetry classes in~$V_5$. We note that each of the endomorphism fields in our list are cyclotomic.

\begin{table}[h!]
  \caption{Frequency of endomorphism fields}
  \label{tab:endo}
  \begin{center}
    \begin{tabular}{rL}
      Frequency  & \text{Defining polynomial} \\
    \hline
60   &  t-1 \\
35   &  t^2 + 1 \\
43   &  t^2 + t + 1 \\
8    &  t^4 - t^2 + 1 \\
7    &  t^6 + t^3 + 1 \\
1    &  t^{12} - t^6 + 1
    \end{tabular}
  \end{center}
\end{table}

\subsection{Methodology}

Once the period matrix of a quartic is approximated, we follow~\cite{lairez-sertoz} to compute the Picard numbers. In order to facilitate the computation of the periods, there are two tricks we used besides the general strategy outlined in Section~\ref{sec:general_framework}.

\subsubsection{Additional simplifications}

The symmetric group $S_4$ acts on the $5$-nomials by permuting the four variables. This is a linear action of the projective space and we can use Section~\ref{sec:linear_translate} to compute period translation matrices at essentially no cost. This connects the elements in each $S_4$-equivalence class.

In order to translate the periods of one polynomial $p$ to another $q$, we need the period matrix of $p$. However, if $p$ is particularly resistant to our computations then we can compute only the first row of the period matrix of $p$ --- reducing the work load by a factor of $21$. With this first row we are still able to compute the Picard number and isomorphism class of $p$. However, $p$ becomes a dead-end; we can no longer use $p$ to compute the periods of another polynomial $q$. 

\subsubsection{Computing isomorphism classes}\label{sec:method_to_compute_iso_classes}

The isomorphism class of a K3 surface depends only on its periods~\cite{looijenga-peters}. In particular, that of the first row of its period matrix. 

Suppose $w_1, w_2 \in \Cc^{22} \simeq \H^2(X,\Cc)$ are periods of two K3s $X_1$ and $X_2$. To detect if $X_1$ and $X_2$ are isomorphic, we need to determine if there exists a constant $c \in \Cc^*$ and an isometry $N \in \Zz^{22 \times 22}$, $N \colon \H_2(X_1,\Zz) \iso \H_2(X_2,\Zz)$, such that
\begin{equation}
  w_1 \cdot N =c  w_2 .
\end{equation}
Using approximations of $w_1$ and $w_2$, this can be translated into a problem of finding short lattice vectors as we describe below. 

The integral relations annihilating $w_1$ and $w_2$ cause a difficulty here. So we first compute the Picard groups $\pic(X_i) \simeq w_i ^{\perp} \subset \Zz^{22}$. If the rank of the Picard groups are distinct then $X_1$ and $X_2$ are not isomorphic.

If $\rho \colonequals \rk \pic(X_1) = \rk \pic(X_2)$, construct $T(X_i) = \pic(X_i)^{\perp} \subset \Zz^{22}$. We can view $w_i$ as an element in $T(X_i) \otimes \Cc \simeq \Cc^{22-\rho}$. Let $v_i \in \Cc^{22-\rho}$ be the new vector corresponding to $w_i$. The surfaces $X_1$ and $X_2$ are isomorphic if and only if there exists $c \in \Cc^*$ and $N' \in \Zz^{(22-\rho) \times (22 - \rho)}$ that satisfy
\begin{equation}
  v_1 \cdot N' = c v_2.
\end{equation}
Following~\cite[\S 2.3]{lairez-sertoz}, we describe how to find such an $N'$ if it exists. That is, if $\langle \cdot,\cdot \rangle$ represents the intersection product on $T(X_2)$ then we need to solve for
\begin{equation}
  \langle v_2,v_1\cdot N' \rangle v_2 = \langle v_2,v_2 \rangle v_1 \cdot N',
\end{equation}
which is linear in $N'$. We now use LLL~\cite{lenstra-82} to find possible integral solutions $N'$. The function {\tt isomorphisms\_of\_k3s} in {\tt PeriodSuite} implements this procedure.

\subsection{Performance on applications}\label{sec:app_performance}

In the end, our goal is to improve the computation time for periods. In Section~\ref{sec:performance} we analyzed the predictive power of the neural network in isolation, but we did not consider its effect on period computation. We describe the influence of the neural network on the task of computing the periods here.

We demonstrate the effect of using a neural network on a small example. The problem here was faced on a larger scale, and faced repeatedly, as we sought to complete the calculations for Table~\ref{tab:pranks}. Given a $5$-nomial $f \in V_5$, we look for $4$-or $6$-nomials $g \in V_4 \cup V_6$ such that the period transition matrix for the edge $(f,g)$ is easy to compute. We used such connections to zig-zag from $V_5$ to $V_4$ or $V_6$ and back to $V_5$ in order to establish new connections between $5$-nomials.

For this example, we chose a random subset $S \subset V_5$ with $100$ elements so that each element in $S$ is from one of the $161$ distinct $S_4$-equivalence classes. For each $f \in S$ we consider the edge set
\begin{equation}
  E_f = \{(f,g) \in V_5 \times V_6 \mid f -g \text{ is a monomial}\}.
\end{equation}
The average size of $E_f$ for $f \in S$ is $29$. 

We compare two methods of exploring the edges $E_f$, one aided by our neural network and one unaided. We used the neural network (ensemble) that was trained on $V_5$ but not on $V_6$ so that there is no extra training time. In particular, \emph{the neural network is faced with a data set for which it has not been trained}; nevertheless it performs well.

For the unaided strategy, we picked $10$ elements from each $E_f$ randomly and tried to compute these edges. For the aided strategy, we sorted $E_f$ using our neural network and picked the top~$10$. The unaided strategy had a $54.4\%$ failure rate as opposed to $33.9\%$ for the aided strategy. Consider the table below that records the frequency of elements in $S$ that had $n$ successful edges for $n \in \{0,\dots,10\}$.

\begin{table}[h!]
  \caption{Frequency of successes}
  \label{tab:success}
  \begin{center}
    \begin{tabular}{r|ccccccccccc}
      \# of connections & 0  & 1 & 2 & 3 & 4  & 5  & 6 & 7 & 8 & 9 & 10 \\ \hline
      Unaided method & 21 & 3 & 5 & 9 & 10 & 16 & 8 & 7 & 3 & 5 & 13 \\
      Aided method & 21 & 0 & 0 & 4 & 4  & 5  & 6 & 4 & 7 & 2 & 47
    \end{tabular}
  \end{center}
\end{table} 

The first column of this table shows that, in both cases, $21$ vertices had $0$ successful edges. This demonstrates the fact that some edges are intrinsically difficult to move away from. Neural networks can sort edges according to difficulty, but they cannot help if every edge is impossible. 

On the other hand, we see that the aided method establishes far more connections to $V_6$. For instance, $47$ vertices in $S$ had all $10$ of their chosen edges successful with the aided method as opposed to $13$ with the unaided method. In practice, this computation would then be repeated for each successful connection, which means that the advantage grows exponentially.

Our computations for the Picard ranks of $V_5$ took a CPU decade. 
The approach presented in this paper allowed us to repeatedly pare down hundreds of thousands of possible edges to a manageable,
but likely to succeed, subset. 
The mini-computation in this section demonstrates the benefit of including a neural ensemble model in the algorithmic pipeline en route to computing period matrices of smooth quartic hypersurfaces.


\bibliography{my.bib}{}
\bibliographystyle{alpha}

\end{document}

%% file: preamble.tex
\usepackage[utf8]{inputenc}
\usepackage[T1]{fontenc} 
\usepackage{microtype}
\usepackage[american]{babel}
\usepackage{pdfpages}
\usepackage{emptypage}

\usepackage{listings}
\usepackage{color}
\usepackage{xcolor}
\usepackage{listings}

\usepackage{caption}
\DeclareCaptionFont{white}{\color{white}}
\DeclareCaptionFormat{listing}{\colorbox{gray}{\parbox{\textwidth}{#1#2#3}}}
\usepackage[normalem]{ulem}

\usepackage{amsmath}
\usepackage{mathtools}
\usepackage{amsthm}
\usepackage{amssymb}
\usepackage{enumitem}

\usepackage{cite}
\usepackage[hyphens]{url}

\usepackage{tikz-cd}
\usepackage{chngcntr}

\usepackage{amsfonts}
\usepackage{yfonts}
\usepackage{mathrsfs}
\usepackage{bbm}
\usepackage[%
  cal=euler,
  scr=rsfso%
]{mathalfa} 

\usetikzlibrary{decorations.pathmorphing}

\DeclareMathOperator{\pic}{Pic}
\DeclareMathOperator{\jac}{Jac}

\DeclareMathOperator{\supp}{supp}
\DeclareMathOperator{\coefs}{coefs}

\DeclareMathOperator{\gl}{GL}

\DeclareMathOperator{\rk}{rk}

\DeclareMathOperator{\res}{res}

\newcommand{\dd}{\mathrm{d}}

\DeclareMathOperator{\vol}{vol}

\DeclareMathOperator{\lt}{lt}

\DeclareMathOperator{\lcm}{lcm}
\DeclareMathOperator*{\argmin}{arg\,min}

\usepackage{array}   
\newcolumntype{M}{>{$}c<{$}} 
\newcolumntype{L}{>{$}l<{$}}
\newcolumntype{R}{>{$}r<{$}}
\usepackage{arydshln}
\usepackage{multirow}

\DeclareMathOperator{\Hom}{\mathscr{H}\kern -2pt om}
\DeclareMathOperator{\Ext}{\mathscr{E}\kern -1.5pt xt}

\newtheorem{theorem}{Theorem}[section]
\numberwithin{theorem}{section} 

\newcommand{\nameofthm}{}

\newtheorem{genericThm}[theorem]{\nameofthm}

{%
  \begin{proof}$ $\newline
  \indent ($\implies$) 
}
{\end{proof}}

\newtheorem*{theorem*}{Theorem}

\theoremstyle{definition}

\newtheorem*{definition*}{Definition}
\newtheorem{definition}[theorem]{Definition}

\newtheorem{remark}[theorem]{Remark}

\newtheorem*{notation*}{Notation}

\newcommand{\nameofenv}{}
\newtheorem*{generic}{\nameofenv}

\newcommand{\nameOfNumEnv}{}
\newtheorem{genericNumbered}[theorem]{\nameOfNumEnv}

\newcommand{\vc}{\mathcal{V}}

\makeatletter
\def\imod#1{\allowbreak\mkern10mu({\operator@font mod}\,\,#1)}
\makeatother

\newcommand{\del}{\partial}

\newcommand{\ppp}{\Pp^3}

\newcommand{\norm}[1]{\lVert #1 \rVert}
\newcommand{\abs}[1]{\lvert #1 \rvert}
\newcommand{\inv}{^{-1}}

\newcommand{\tos}{\twoheadrightarrow}

\newcommand{\isoto}{\overset{\sim}{\to}}
\newcommand{\iso}{\overset{\sim}{\to}}

\newcommand{\Cc}{\mathbbm{C}}

\newcommand{\Nn}{\mathbbm{N}}
\newcommand{\Qq}{\mathbbm{Q}}
\newcommand{\Pp}{\mathbbm{P}}
\newcommand{\Rr}{\mathbbm{R}}

\newcommand{\Zz}{\mathbbm{Z}}

\renewcommand{\H}{\mathrm{H}}

\newcommand{\cc}{\mathcal{C}}
\newcommand{\cd}{\mathcal{D}}

\newcommand{\ci}{\mathcal{I}}

\newcommand{\cn}{\mathcal{N}}

\newcommand{\cp}{\mathcal{P}}

\newcommand{\cs}{\mathcal{S}}
\newcommand{\ct}{\mathcal{T}}

\newcommand{\cx}{\mathcal{X}}

\newcommand{\prim}{0}

\newcommand{\K}{\sf{K}}

\newcommand{\ie}{i.e.~}

